\documentclass[11pt]{article}
\usepackage{pifont}
\usepackage{booktabs,graphicx}

\usepackage[final]{acl}
\usepackage{times}
\usepackage{latexsym}

\usepackage[T1]{fontenc}

\usepackage[utf8]{inputenc}

\usepackage{float}
\usepackage{microtype}

\usepackage{inconsolata}
\usepackage{algorithm}
\usepackage{algpseudocode}
\usepackage{amsmath,amssymb}
\usepackage{booktabs}
\usepackage{enumitem}
\usepackage{tabularx,booktabs,makecell}
\usepackage{graphicx}
\usepackage{subcaption}
\usepackage{float}
\newcommand{\yesmark}{\mbox{\ding{51}}}
\newcommand{\nomark}{\mbox{\ding{55}}}

\newcommand{\datasetmain}{BlogText}

\newcommand{\meanstd}[2]{%
  \makecell[c]{#1\\[-1pt]{\tiny $\pm\,#2$}}%
}

\title{FAVoR: Measuring and Mitigating Author-Style Homogenization in Federated Personalized Generation}

\author{
  \textbf{Lu Han}\textsuperscript{*},
  \textbf{Jingyao Zhang}\textsuperscript{*},
  \textbf{Katy Ilonka Gero},
  \textbf{Nguyen H. Tran}
  \\
  School of Computer Science, The University of Sydney
  \\
  Sydney, NSW 2006, Australia
  \\
  \small{
    \textsuperscript{*}\textbf{Corresponding authors:}
    \texttt{\{lhan0123,jzha0544\}@sydney.edu.au}
  }
}

\begin{document}
\maketitle
\begin{abstract}
Large language models are increasingly used as personalized writing assistants, but adapting a model across many authors can compromise individual writing style by pulling author-specific signals toward a shared register.
Federated parameter-efficient fine-tuning (PEFT) offers a data-local setting for this multi-author adaptation problem: clients keep author text local while sharing compact adapter updates.
However, we show that standard aggregation can preserve continuation utility while making different authors' generations less distinguishable in style space, a failure mode we define as \emph{author-style homogenization}.
We evaluate author-style retention with Angular Style Classification Encoder (ASCE)-based diagnostics on our main BlogText benchmark and ASCE-independent external authorship verification.
Using this protocol, we find that common federated PEFT baselines can preserve semantic utility while averaging out author-specific signals.
To address this homogenization, we instantiate FAVoR (Federated Authorial Voice Retention), an author-style residual mechanism for federated PEFT.
FAVoR uses a shared--private adapter design: clients upload shared-adapter updates while retaining author-specific residual corrections locally.
Across BlogText and external Mythos-Reddit validation, FAVoR improves author-style retention over standard and personalized federated PEFT baselines.
These gains come with small continuation-utility trade-offs and are supported by component ablations, external verification, and cold-start transfer. 
\end{abstract}

\section{Introduction}
\label{sec:introduction}

Personalized writing support often asks a large language model (LLM) to continue unseen prompts while preserving an author's characteristic writing style.
In collaborative adaptation, however, improvements in general continuation competence can reduce the stylistic distinctions that personalization is meant to preserve.
We call this failure mode \textbf{author-style homogenization}: continuations remain fluent and semantically plausible, but lose author-specific stylistic signals.

This problem is especially sharp under author-level non-identically distributed (non-IID) data.
Clients differ in lexical choice, syntax, pacing, rhetorical structure, and register, not merely in sample distribution.
Na\"ively pooling author data, or averaging all parameter-efficient fine-tuning (PEFT) updates in federated learning (FL), can mix author-specific correction directions with unrelated or opposing directions from other authors, shifting generations toward a shared register.
Keeping raw text local is therefore insufficient by itself \citep{lin-etal-2022-fednlp}; standard FL still needs a mechanism for deciding which adaptation directions should be shared and which should remain client-specific.

We study this issue in \textbf{personalized held-out continuation}: each client corresponds to one author, and the model must continue held-out prompts in that author's style.
This setting makes homogenization directly testable, because collaboration should improve continuation quality without erasing cross-author stylistic distinctions, and evaluation must separate author signal from topical content.

\begin{figure}[t]
    \centering
    \vspace{-6pt}
    \includegraphics[
        width=0.92\linewidth,
        trim=0 4pt 0 6pt,
        clip
    ]{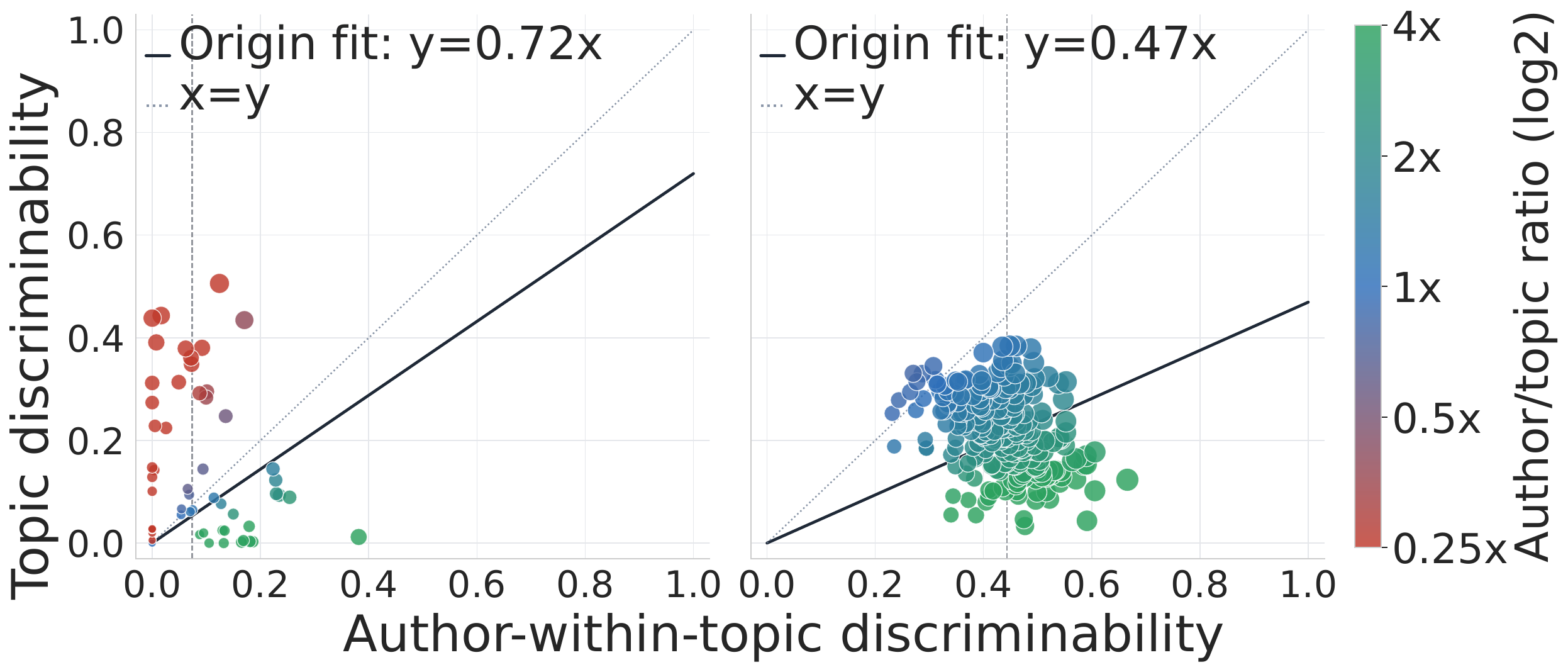}
    \vspace{-6pt}
    \caption{Coordinate-level author/topic discriminability. Each point is one scalar coordinate: a hand-crafted stylometric feature on the left or an ASCE embedding dimension on the right. Lower-right and greener points carry more author-than-topic signal; marker size scales with distance from the origin. ASCE shifts the distribution toward stronger author signal and a lower topic-on-author origin-fit slope.}
    \label{fig:signature_space_discriminability}
    \vspace{-6pt}
\end{figure}

We introduce an \textbf{author-style retention evaluation protocol} based on an Author-Style Signature Space.
For BlogText, we instantiate this space with an author-disjoint Angular Style Classification Encoder (ASCE).
To test whether this space captures the author signal rather than the topic alone, we use the topic as a coarse content proxy and the author labels as an operational proxy for author style.
Figure~\ref{fig:signature_space_discriminability} shows that the learned signature space increases author-within-topic separability and reduces the topic--author origin-fit slope from $0.72$ to $0.47$ relative to raw computational-linguistic features.

To mitigate homogenization, we introduce FAVoR, an author-style residual mechanism on a shared--private federated PEFT pattern.
The shared adapter carries transferable continuation updates, while client-local residual adapters retain author-specific corrections outside aggregation. 
We summarize the contributions of this work as follows:
\begin{itemize}
    \setlength{\itemsep}{2pt}
    \setlength{\parskip}{0pt}
    \setlength{\parsep}{0pt}
    \setlength{\topsep}{2pt}

    \item We define \textbf{author-style homogenization} as a utility-preserving loss of cross-author stylistic distinguishability in federated held-out continuation.

    \item We introduce an \textbf{author-style retention evaluation protocol} combining author-disjoint ASCE diagnostics, ASCE-independent gen-to-source verification, and source-space topic controls.

    \item We propose \textbf{FAVoR}, an author-style residual mechanism for federated PEFT. FAVoR separates shared continuation competence from client-local style residuals, optimizes the residual with local author-style targets, and keeps it outside aggregation.

    \item Across BlogText and Mythos-Reddit, we show that FAVoR improves author-style retention over standard and personalized federated PEFT baselines while maintaining a favorable retention--utility trade-off.
\end{itemize}

\section{Related Work}

\paragraph{Federated Personalization and PEFT for LLMs.}
Federated learning enables decentralized model averaging without sharing raw client data, but statistical heterogeneity can cause a single global model to underfit client-specific behavior \citep{mcmahan2017communication,lin-etal-2022-fednlp}.
Personalized FL addresses this issue through client-specific objectives, proximal regularization, or shared/client-specific adaptation structures, including pFedMe, Ditto, FedProx, and recent federated adapter variants \citep{NEURIPS2020_f4f1f13c,pmlr-v139-li21h,li-etal-2020-fedprox,qi2024fdlorapersonalizedfederatedlearning,NEURIPS2024_45a30141}.
Recent federated LLM work combines FL with parameter-efficient adaptation, commonly through low-rank adapters (LoRA), to reduce communication and memory cost \citep{hu2022lora,wang-etal-2024-flora,bai-etal-2024-flexlora}.
FAVoR uses this global--local adapter pattern as a communication design rather than claiming it as the architectural novelty. Its contribution is to study author-style homogenization in federated held-out continuation and to define the local component as an author-style residual, trained with local author targets and kept outside aggregation.

\paragraph{Authorship-Conditioned Generation and Evaluation.}
Mythos constructs author profiles from multiple writings and evaluates whether the generated stories resemble held-out author-written stories \citep{ashok-kumar-etal-2025-whose}, motivating author-conditioned generation rather than federated collaboration among private authors.
Style transfer and rewriting are related but distinct: formality-transfer benchmarks transform a given input while preserving content \citep{rao-tetreault-2018-dear}, whereas held-out continuation requires open-ended generation that preserves coherence, distributional fit, and author-specific style.
We use MAUVE and BERTScore as continuation-utility metrics \citep{JMLR:v24:23-0023,zhang-etal-2020-bertscore}, but additionally evaluate author-style retention, inter-author separability, and collapse because fluent continuations can still converge to a shared register.
Human--AI writing studies report related homogenization effects under model assistance \citep{padmakumar2024contentdiversity,agarwal2025aihomogenize}.

\paragraph{Style and Authorship Representation Learning.}

Authorship encoders such as LUAR and STAR learn representations for authorship verification and cross-domain transfer, while also revealing domain entanglement in learned author features \citep{rivera-soto-etal-2021-learning,huertas-tato-etal-2024-star}.
Style-specific embedding work shows that naturally occurring contrastive examples can leak topic; StyleDistance addresses this with content-controlled style comparisons \citep{patel-etal-2025-styledistance}.
Our normalized author/style space also connects to ArcFace-style metric learning for identity separation \citep{deng2019arcface}.
We build on this line of work by using author/style representations both as evaluation spaces and as local alignment targets.

\section{Method}
\label{sec:method}
\setlength{\abovedisplayskip}{3pt plus 1pt minus 1pt}
\setlength{\belowdisplayskip}{3pt plus 1pt minus 1pt}
\setlength{\abovedisplayshortskip}{2pt plus 1pt minus 1pt}
\setlength{\belowdisplayshortskip}{2pt plus 1pt minus 1pt}
\setlength{\textfloatsep}{6pt plus 2pt minus 2pt}
\setlength{\floatsep}{6pt plus 2pt minus 2pt}
\setlength{\intextsep}{6pt plus 2pt minus 2pt}
\setlength{\dbltextfloatsep}{6pt plus 2pt minus 2pt}
\setlength{\dblfloatsep}{6pt plus 2pt minus 2pt}
\setlength{\abovecaptionskip}{3pt plus 1pt minus 1pt}
\setlength{\belowcaptionskip}{0pt}

\begin{figure*}[t]
    \centering
    \includegraphics[width=0.97\textwidth]{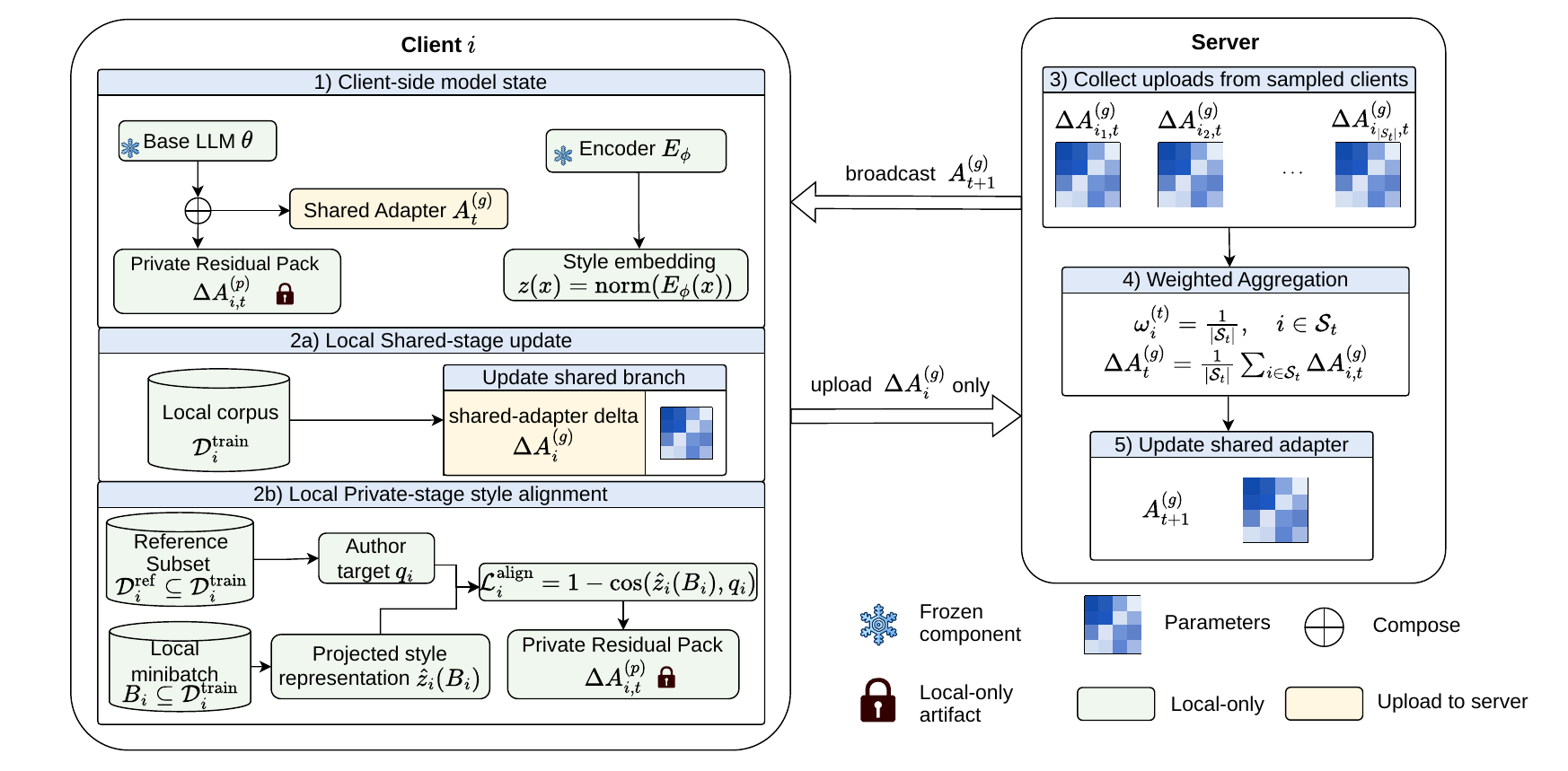}
    \caption{FAVoR workflow. Clients upload only shared-adapter deltas for aggregation; author targets, hidden states, and private residual packs remain local. Private updates use task loss plus style alignment.}
    \label{fig:favor_overview}
\end{figure*}

We study data-local federated personalized held-out continuation under \emph{author-non-IID} data.
There are $N$ clients, where each client $i \in \{1,\dots,N\}$ corresponds to one author with a local training corpus $\mathcal{D}_i^{\mathrm{train}}$.
The goal is to improve shared continuation competence across clients while preserving author-specific style, so that collaboration does not homogenize outputs across authors.
Each client also keeps a local author-reference set $\mathcal{D}_i^{\mathrm{ref}}\subseteq\mathcal{D}_i^{\mathrm{train}}$ for style estimation.
FAVoR imposes a shared--private aggregation boundary: the shared adapter is the only federated component, whereas private residual packs remain local and are never aggregated.
Figure~\ref{fig:favor_overview} gives the workflow; Algorithm~\ref{alg:favor_main} summarizes one round.

\subsection{Mechanism Constraint: Shared Competence and Private Style}
\label{subsec:shared_private}

FAVoR stores the private adapter as a residual pack relative to the shared endpoint (Eq.~\eqref{eq:residual_pack}) and trains it against a local author target.
Together with this residual structure, the shared--private boundary separates shared continuation competence from author-specific stylistic variation, limiting author-style homogenization.

Starting from a frozen base language model with parameters $\theta$, FAVoR uses two PEFT artifacts with asymmetric communication roles:
\begin{itemize}[itemsep=1pt, topsep=2pt, parsep=0pt, partopsep=0pt]
    \item a \textbf{shared adapter} $A^{(g)}$, which is communicated and aggregated across clients to learn transferable continuation competence; and
    \item a \textbf{private residual pack} $\Delta A_{i,\tau_i}^{(p)}$, where $\tau_i$ is client $i$'s latest private-update round; it remains local and stores the author-specific correction from the shared endpoint.
\end{itemize}

Throughout, $A$ denotes any serialized LoRA adapter-shaped state---including shared, personalized, and residual/delta states---i.e., a set of factor tensors $(U_\ell,V_\ell)$ for each layer $\ell$, not the LoRA factor conventionally called $A$. With LoRA rank $r$ and scaling $\alpha$, the effective per-layer update is $\Delta W_\ell(A) = (\alpha/r)V_\ell U_\ell$.
The personalized model for client $i$ at its latest private update $\tau_i$ is
\begin{equation}
p_i(y \mid x)
=
\mathrm{LM}\!\left(y \mid x;\ \theta \oplus \left(A_{i,\tau_i}^{(g)} + \Delta A_{i,\tau_i}^{(p)}\right)\right),
\label{eq:favor_model}
\end{equation}
where $\mathrm{LM}(y\mid x;\cdot)$ denotes the autoregressive likelihood of response tokens $y$ conditioned on $x$, $\oplus$ attaches an adapter to the frozen base, and $A^{(g)}+\Delta A^{(p)}$ denotes factor-wise reconstruction of the LoRA adapter state.
The private residual pack is defined at the LoRA-factor level, relative to the client's shared endpoint:
\begin{equation}
\begin{aligned}
\Delta A_{i,\tau_i}^{(p)}
&=
\left\{\Delta U_{i,\tau_i,\ell}^{(p)},\Delta V_{i,\tau_i,\ell}^{(p)}\right\}_{\ell},\\
\Delta U_{i,\tau_i,\ell}^{(p)}
&=
U_{i,\tau_i,\ell}^{(p)} - U_{i,\tau_i,\ell}^{(g)},\\
\Delta V_{i,\tau_i,\ell}^{(p)}
&=
V_{i,\tau_i,\ell}^{(p)} - V_{i,\tau_i,\ell}^{(g)} .
\end{aligned}
\label{eq:residual_pack}
\end{equation}
At inference, reconstructing the LoRA factors as $U^{(g)}_{\ell}+\Delta U^{(p)}_{\ell}$ and $V^{(g)}_{\ell}+\Delta V^{(p)}_{\ell}$ exactly recovers the locally trained rank-$r$ adapter $A_{i,\tau_i}^{(p)}$, so the personalized rank is preserved by construction.
Thus, Eq.~\eqref{eq:residual_pack} defines how the personalized adapter is stored locally and reconstructed, rather than assuming a linear decomposition of the effective LoRA weight update.
We do not assume effective-weight additivity ($\Delta W_\ell(A^{(g)}+\Delta A^{(p)})\!\neq\!\Delta W_\ell(A^{(g)})+\Delta W_\ell(\Delta A^{(p)})$ in general); cross terms remain local during reconstruction.
The operative shared--private boundary lies in the aggregation policy and the shared-endpoint coupling of Sec.~\ref{subsec:local_updates}: only $A^{(g)}$ updates are uploaded, so aggregation never averages residual packs across authors.

\subsection{Angular Style Classification Encoder}
\label{subsec:asce}

To make author preservation trainable and measurable, FAVoR uses a fixed style encoder, which we call the \textbf{Angular Style Classification Encoder} (ASCE).
ASCE maps a text $x$ to a normalized style embedding:
\begin{equation}
z(x) = \operatorname{norm}\!\left(E_{\phi}(x)\right).
\label{eq:style_embedding}
\end{equation}
Here $E_\phi$ is the ASCE text encoder and $\operatorname{norm}(v)=v/\|v\|_2$.

ASCE is trained to make texts from the same author compact and texts from different authors separable.
Concretely, we train it with an angular-margin authorship classification objective, inspired by angular-margin metric learning in face recognition \citep{deng2019arcface}.
Let $K_{\mathrm{ASCE}}$ be the number of ASCE-training authors, $a$ the author index of text $x$, and $\{w_k\}_{k=1}^{K_{\mathrm{ASCE}}}$ the learned normalized classifier prototypes, with $w_a$ selecting the positive-author prototype. These prototypes are discarded with the classifier head after ASCE training.
We first define the positive-author angle $\theta_a(x)$ and logits $s_a(x)$ and $s_k(x)$ as
\begin{equation}
\begin{aligned}
\theta_a(x) &= \arccos\!\left(w_a^\top z(x)\right),\\
s_a(x) &= \gamma \cos\!\left(\theta_a(x)+m_s\right),\\
s_k(x) &= \gamma w_k^\top z(x), \quad k\neq a .
\end{aligned}
\label{eq:asce_logits}
\end{equation}

The ASCE training loss is the cross-entropy loss over these logits:
\begin{equation}
\begin{aligned}
\mathcal{L}^{\mathrm{ASCE}}(x,a)
&=
-s_a(x) + \log Z_a(x)\\
\text{where} \quad  
Z_a(x)
&=
\exp(s_a(x)) + \sum_{k\neq a}\exp(s_k(x)). 
\end{aligned}
\label{eq:asce_loss}
\end{equation}
Here $s_a$ is the margin-adjusted target-author logit, $s_k$ are non-target logits, $Z_a$ is the softmax normalizer, $m_s$ is the angular margin, and $\gamma$ rescales normalized cosine logits.
After ASCE training, the classification head is discarded and the encoder $E_{\phi}$ is fixed.

\paragraph{Training-time status.}
ASCE is trained once before the federated procedure, and its encoder $E_{\phi}$ is then frozen.
During FL, clients use this fixed encoder only as a local style-measurement module: each client computes its author target from its local reference set and computes private-stage alignment signals from local minibatches on the device.

For each client $i$, FAVoR constructs the local author target in Eq.~\eqref{eq:author_target} from the deterministic local author-reference set $\mathcal{D}_i^{\mathrm{ref}}$:
\begin{equation}
q_i =
\operatorname{norm}\!\left(
\frac{1}{|\mathcal{D}_i^{\mathrm{ref}}|}
\sum_{x \in \mathcal{D}_i^{\mathrm{ref}}}
z(x)
\right).
\label{eq:author_target}
\end{equation}
This target remains local and serves as the client's private style anchor in the private-stage alignment loss in Eq.~\eqref{eq:style_alignment_loss}.

\subsection{Local Shared and Private Updates}
\label{subsec:local_updates}

Each sampled client performs a shared-stage update followed by a private-stage update, as shown in Algorithm~\ref{alg:favor_main}, lines~\ref{line:favor_shared_update}--\ref{line:favor_private_update}.

For a local minibatch $B_i$, both the shared-stage update and the private-stage update use the same task loss:
\begin{equation}
\mathcal{L}^{\mathrm{task}}_i(B_i)
=
-\frac{1}{|B_i|}
\sum_{(x,y)\in B_i}
\log p_i(y \mid x).
\label{eq:task_loss}
\end{equation}
In Eq.~\eqref{eq:task_loss}, $p_i$ is evaluated under the adapter being updated: the local shared copy (the client's copy of the broadcast shared adapter) in the shared stage and the composed personalized model in Eq.~\eqref{eq:favor_model} in the private stage.

The local shared copy is trained with the objective in Eq.~\eqref{eq:shared_objective}, combining the task loss with a proximal stabilization term around the current server adapter:
\begin{equation}
\mathcal{L}^{\mathrm{shared}}_i
=
\mathcal{L}^{\mathrm{task}}_i
+
\lambda_{\mathrm{prox}}
\left\|A_i^{(g)} - A_t^{(g)}\right\|_2^2.
\label{eq:shared_objective}
\end{equation}
This term discourages local shared updates from drifting too far from the round-$t$ global adapter; the norm sums squared Frobenius norms over LoRA factor tensors.
Because the shared-stage update produces the communicated adapter delta, the explicit style-alignment signal is not applied here; it is reserved for the local private-stage update.

The private-stage update uses an additional style-alignment objective.
Rather than sampling generations during training, FAVoR uses teacher-forced hidden states from the personalized model.
Let $H_i(B_i) \in \mathbb{R}^{|B_i| \times L \times d}$ denote the final-layer hidden states produced on minibatch $B_i$. Let $M_i(B_i) \in \{0,1\}^{|B_i| \times L}$ be the supervised-response-token mask, set to $1$ at positions whose label is not the ignore index and $0$ otherwise (i.e., prompt and padding positions are excluded). The pooled style state in Eq.~\eqref{eq:pooled_hidden} is the mask-weighted mean of $H_i$ over response tokens:
\begin{equation}
h_i(B_i)
=
\frac{\sum_{b,\ell} M_{i,b,\ell}\, H_{i,b,\ell}}
{\sum_{b,\ell} M_{i,b,\ell}}.
\label{eq:pooled_hidden}
\end{equation}
The projected style representation in Eq.~\eqref{eq:projected_style} is then obtained with a lightweight projection head $P_{\psi}$:
\begin{equation}
\hat{z}_i(B_i)
=
\operatorname{norm}\!\left(P_{\psi}(h_i(B_i))\right).
\label{eq:projected_style}
\end{equation}
$P_\psi$ is a lightweight 2-layer MLP from the LM hidden size to the ASCE embedding dimension. It is randomly initialized at the start of each client's private stage, optimized jointly with the private adapter under $\mathcal{L}^{\mathrm{align}}_i$, and discarded afterwards. It is therefore per-client and per-round, is never uploaded or aggregated, and is not used at inference; because it is discarded after training, only gradients propagated through it to the private adapter can have a persistent effect.

The private-stage style-alignment loss in Eq.~\eqref{eq:style_alignment_loss} measures cosine distance to the local author target from Eq.~\eqref{eq:author_target}:
\begin{equation}
\mathcal{L}^{\mathrm{align}}_i
=
1 -
\cos\!\left(\hat{z}_i(B_i), q_i\right).
\label{eq:style_alignment_loss}
\end{equation}

The private-stage objective in Eq.~\eqref{eq:private_objective} combines the task loss from Eq.~\eqref{eq:task_loss} with the style-alignment loss from Eq.~\eqref{eq:style_alignment_loss}:
\begin{equation}
\mathcal{L}^{\mathrm{private}}_i
=
\mathcal{L}^{\mathrm{task}}_i
+
\lambda_{\mathrm{align}} \kappa_t
\mathcal{L}^{\mathrm{align}}_i,
\label{eq:private_objective}
\end{equation}
where $\lambda_{\mathrm{align}}$ controls the strength of style alignment and $\kappa_t \in [0,1]$ is a warmup coefficient. In our experiments, $\kappa_t$ linearly ramps from $0$ to $1$ over the first $5\%$ of private-stage steps in each round and is held at $1$ thereafter. We set $\lambda_{\mathrm{align}} = 0.3$.
In the private stage, forward passes use the composed personalized model in Eq.~\eqref{eq:favor_model}; gradients are stopped through the shared endpoint $A_{i,t}^{(g)}$ and applied only to the private adapter/residual parameters.
This isolates the author-specific correction as a residual relative to the shared endpoint rather than as another contribution to the public aggregation space.
The resulting personalized adapter is stored as the private residual pack via Eq.~\eqref{eq:residual_pack}.

\subsection{Federated Aggregation}
\label{subsec:federated_aggregation}

At communication round $t$, the server samples a subset of clients $\mathcal{S}_t$ and broadcasts the current shared adapter $A_t^{(g)}$.
Each sampled client performs its local shared update and uploads only the shared-adapter delta in Eq.~\eqref{eq:shared_delta}:
\begin{equation}
\Delta A_{i,t}^{(g)}
=
A_{i,t}^{(g)} - A_t^{(g)}.
\label{eq:shared_delta}
\end{equation}
Only $\Delta A_{i,t}^{(g)}$ is uploaded; all residuals, author targets, hidden states, references, and style-control signals remain local.

The server aggregates sampled client deltas using Eq.~\eqref{eq:shared_aggregation_delta}:
\begin{equation}
\Delta A_t^{(g)}
=
\sum_{i \in \mathcal{S}_t}
\omega_i^{(t)}
\Delta A_{i,t}^{(g)},
\qquad
\sum_{i \in \mathcal{S}_t}\omega_i^{(t)} = 1.
\label{eq:shared_aggregation_delta}
\end{equation}
Because each client represents one author, we use uniform sampled-client weighting in the main experiments, i.e.,
$\omega_i^{(t)} = 1/|\mathcal{S}_t|$.
The shared adapter is then updated according to Eq.~\eqref{eq:shared_adapter_update}:
\begin{equation}
A_{t+1}^{(g)}
=
A_t^{(g)}
+
\eta_g
\Delta A_t^{(g)},
\label{eq:shared_adapter_update}
\end{equation}
where $\eta_g$ is the server learning rate.

\begin{algorithm}[t]
\caption{FAVoR Round Summary}
\label{alg:favor_main}
\vspace{-3pt} 
\footnotesize
\begin{algorithmic}[1]
\Require Current shared adapter $A_t^{(g)}$; sampled clients $\mathcal{S}_t$; fixed ASCE encoder $E_{\phi}$
\State Server broadcasts $A_t^{(g)}$ to clients in $\mathcal{S}_t$
\ForAll{clients $i \in \mathcal{S}_t$ in parallel}
    \State \label{line:favor_shared_update} Train a local shared copy with $\mathcal{L}^{\mathrm{shared}}_i$, yielding $A_{i,t}^{(g)}$
    \State \label{line:favor_private_update} Train the local private adapter from $A_{i,t}^{(g)}$ with $\mathcal{L}^{\mathrm{private}}_i$
    \State Store $\{A_{i,t}^{(g)},\, \Delta A_{i,t}^{(p)}\}$ locally and upload only $\Delta A_{i,t}^{(g)}$
\EndFor
\State Server aggregates uploaded shared deltas and updates $A_{t+1}^{(g)}$
\Statex \textit{Full pseudocode in Algorithm~\ref{alg:favor_full}.}
\end{algorithmic}
\end{algorithm}

\paragraph{Data-locality boundary.}
FAVoR follows the standard federated-learning data-locality boundary (Figure~\ref{fig:favor_overview}, Algorithms~\ref{alg:favor_main}--\ref{alg:favor_full}).
The server receives and aggregates only shared-adapter deltas $\Delta A_{i,t}^{(g)}$.
Client training corpora $\mathcal{D}_i^{\mathrm{train}}$, local author-reference sets $\mathcal{D}_i^{\mathrm{ref}}$, author targets $q_i$, private residual packs $\Delta A_{i,t}^{(p)}$, style-alignment hidden states, and local style-control signals remain on device and outside aggregation.
This is a data-locality claim rather than a formal privacy guarantee: shared-adapter deltas may still encode memorized text or author-identifying information, and we do not implement secure aggregation, differential privacy, or update-level leakage auditing.
The fixed ASCE encoder in Eq.~\eqref{eq:style_embedding} is a centralized pre-training artifact, frozen during FL. For BlogText, its training authors are disjoint from the FL roster, and during FL, client data are used only to compute local style-control signals.

\paragraph{Inference.}
At inference, client $i$ composes the frozen base model with $A_{i,\tau_i}^{(g)}+\Delta A_{i,\tau_i}^{(p)}$ as in Eq.~\eqref{eq:favor_model}.
Because the residual is tied to the shared endpoint that produced it, unsampled clients retain their latest personalized endpoint rather than recomposing that residual with a later global adapter.

\setlength{\abovedisplayskip}{10pt plus 2pt minus 5pt}
\setlength{\belowdisplayskip}{10pt plus 2pt minus 5pt}
\setlength{\abovedisplayshortskip}{0pt plus 3pt}
\setlength{\belowdisplayshortskip}{6pt plus 3pt minus 3pt}
\setlength{\textfloatsep}{20pt plus 2pt minus 4pt}
\setlength{\floatsep}{12pt plus 2pt minus 2pt}
\setlength{\intextsep}{12pt plus 2pt minus 2pt}
\setlength{\dbltextfloatsep}{20pt plus 2pt minus 4pt}
\setlength{\dblfloatsep}{12pt plus 2pt minus 2pt}
\setlength{\abovecaptionskip}{10pt}
\setlength{\belowcaptionskip}{0pt}
\section{Experiment}
\label{sec:experiments}

\paragraph{Data preprocessing and metrics.}
Our main benchmark, BlogText, is a filtered held-out continuation split of the Blog Authorship Corpus \citep{schler-etal-2006-blogging} based on Tatman's Kaggle redistribution \citep{tatman-blog-authorship-corpus-kaggle}.
We retain authors with at least 30K tokenizer tokens, split each author's documents into 80/10/10 train/validation/test partitions, and sample a style-diverse 50-author roster. Each federated client is one author, so raw author text remains client-local.
Evaluation prompts are 96-token prefixes of held-out posts (192 tokens on Mythos-Reddit), and generations are capped at 220 new tokens.
We filter predefined degenerate outputs: empty or short generations with fewer than 20 whitespace-delimited tokens, and repetition-collapsed generations with a distinct-bigram ratio below $0.15$.  Appendix~\ref{subsec:ablation_significance_audit} reports the coverage check.
External validation uses 50 Mythos-Reddit r/WritingPrompts authors and 300 balanced held-out continuations \citep{ashok-kumar-etal-2025-whose,fan-etal-2018-hierarchical}.

We evaluate author-style retention using ASCE-space separability, geometry, and collapse metrics, plus external gen$\to$source verification with StyleDistance, MiniLM, and stylometric authorship features \citep{patel-etal-2025-styledistance,sentence-transformers-minilm-l6-v2,stamatatos2009survey}.
Unless otherwise noted, result tables report author accuracy (Acc.), macro-F1, receiver-operating-characteristic area under the curve (AUC), and equal error rate (EER).
Semantic similarity is abbreviated as SemSim.
We run a small human two-alternative forced-choice (2AFC) audit in which annotators choose which of two author profiles better matches a candidate continuation, used as a perceptual sanity check rather than a full preference or authenticity evaluation (see Appendix~\ref{subsec:human_2afc_audit}).

\paragraph{Experimental setup.}
\label{subsec:experiment_protocol}
\label{subsec:experiment_methods}
\label{subsec:experiment_metrics}
The experiments test whether collaborative PEFT homogenizes author style, whether FAVoR mitigates this effect without degrading continuation quality, and whether its author-style-retention advantage over collaborative baselines transfers beyond BlogText. The main BlogText evaluation uses three seeds on a frozen 50-client medium non-IID long-tail split (target/realized Gini: $0.25/0.196$; $904{,}378$ allocated word units defined by a fixed English-word regular expression; FL schedule: $(R,C,\bar{p})=(6,12,1.44)$).

Table~\ref{tab:main_federated_results} focuses on collaborative federated methods (i.e., FedAvg, FedProx, pFedMe, Ditto, FedDPA, and FAVoR) while pooled PEFT is reported as a centralized non-federated reference in Table~\ref{tab:phase1-main-external-verification}.
FedDPA is included as the closest-family global--local adapter baseline.
Local-only adaptation is treated as a strong non-collaborative style-retention reference, so base-only and local-only references are reserved for diagnostic and appendix tables and excluded from Table~\ref{tab:main_federated_results}.

We ablate ASCE alignment and the shared--private residual decomposition.
Appendices~\ref{subsec:experiment_prompting_baselines}--\ref{subsec:experiment_significance} report prompting/retrieval-style baselines, source-space controls, sensitivity checks, external-verifier breakdowns, communication costs, and full statistical tests.
Training objectives and local losses are specified in Section~\ref{sec:method}. Final backbone, adapter, batching, learning-rate, communication, storage, and runtime settings are reported in Appendix~\ref{subsec:experiment_efficiency}.

\paragraph{ASCE instantiation and evaluator boundary.}
For BlogText, the ASCE encoder used for local style alignment and ASCE-space author-style evaluation is trained once before FL on 50 BlogText authors disjoint from the 50-author FL roster. It does not observe FL training texts, held-out prompts, generated continuations, or any main-roster author identity. We discard its classifier head and use the frozen encoder to define style targets and evaluation embedding. Appendix~\ref{subsec:experiment_asce_leakage_free_audit} gives the training and evaluation details. ASCE-independent verifiers provide a complementary check in Table~\ref{tab:phase1-main-external-verification} and Appendix~\ref{subsec:experiment_external_verifiers}.

\subsection{Signature-Space Validity}
\label{subsec:experiment_signature_space}
Before using ASCE for author-style retention measurement, we test whether the author-style signal is recoverable under topic controls.
Table~\ref{tab:signature-space-main} reports the compact source-space diagnostic against RoBERTa, StyleDistance, LUAR, and STAR baselines \citep{liu-etal-2019-roberta,patel-etal-2025-styledistance,rivera-soto-etal-2021-learning,huertas-tato-etal-2024-star}. Appendix~\ref{subsec:experiment_source_controls} gives the expanded version.

\begin{table}[H]
\centering
\footnotesize
\setlength{\tabcolsep}{3.2pt}
\renewcommand{\arraystretch}{1.05}
\begin{tabular}{lcccc}
\toprule
Representation & Acc. $\uparrow$ & F1 $\uparrow$ &
AUC $\uparrow$ & EER $\downarrow$ \\
\midrule
RoBERTa & 0.635 & 0.677 & 0.629 & 0.421 \\
StyleDist. & 0.628 & 0.660 & 0.716 & 0.344 \\
LUAR & 0.633 & 0.677 & 0.797 & 0.277 \\
STAR & \underline{0.648} & \textbf{0.715} & \underline{0.856} & \underline{0.229} \\
ASCE & \textbf{0.651} & \underline{0.692} & \textbf{0.863} & \textbf{0.211} \\
\bottomrule
\end{tabular}
\caption{Compact source-space representation diagnostics. Coordinate-level author/topic discriminability is reported as an appendix diagnostic in Table~\ref{tab:signature-space-native}.}
\label{tab:signature-space-main}
\end{table}

\subsection{Main Results and Ablations}
\label{subsec:experiment_main_blog}
\label{subsec:experiment_ablations}
\paragraph{Main results.}

\begin{table*}[h!]
\centering
\footnotesize
\setlength{\tabcolsep}{3.5pt}
\renewcommand{\arraystretch}{1.08}
\begin{tabular}{llcccccc}
\toprule
Dataset & Method & \makecell{Author\\Acc. $\uparrow$} & \makecell{Macro-\\F1 $\uparrow$} & \makecell{Verif.\\AUC $\uparrow$} & \makecell{EER\\$\downarrow$} & \makecell{BERTScore\\$\uparrow$} & \makecell{SemSim\\$\uparrow$} \\
\midrule
\datasetmain & FedAvg  & 0.199 & 0.164 & 0.680 & 0.371 & \textbf{0.819} & 0.359 \\
 & FedProx & 0.174 & 0.161 & 0.674 & 0.365 & 0.817 & 0.353 \\
 & pFedMe  & 0.271 & 0.252 & 0.739 & 0.321 & 0.817 & 0.372 \\
 & Ditto   & 0.278 & 0.255 & \underline{0.763} & \underline{0.297} & 0.815 & 0.366 \\
 & FedDPA\textsuperscript{\dag} & \underline{0.294} & \underline{0.271} & 0.749 & 0.313 & \underline{0.818} & \underline{0.376} \\
 & FAVoR   & \textbf{0.403} & \textbf{0.365} & \textbf{0.824} & \textbf{0.245} & 0.815 & \textbf{0.380} \\
\midrule
Mythos-Reddit & FedAvg & 0.134 & 0.115 & 0.658 & 0.386 & 0.823 & 0.482 \\
 & FedProx & 0.141 & 0.113 & 0.663 & 0.377 & 0.823 & 0.478 \\
 & pFedMe & 0.324 & 0.290 & 0.762 & 0.305 & 0.823 & 0.477 \\
 & Ditto & \underline{0.392} & \underline{0.359} & \underline{0.818} & \underline{0.253} & \underline{0.825} & 0.487 \\
 & FedDPA\textsuperscript{\dag} & 0.357 & 0.317 & 0.803 & 0.268 & 0.825 & \underline{0.488} \\
 & FAVoR & \textbf{0.561} & \textbf{0.521} & \textbf{0.900} & \textbf{0.175} & \textbf{0.826} & \textbf{0.494} \\
\bottomrule
\end{tabular}
\caption{Main federated held-out continuation results. \textsuperscript{\dag}FedDPA is trained with the dynamic-gate setup; reported continuations use the exported per-client mean validation gate, audited in Appendix~\ref{subsec:feddpa_gate_sensitivity}.}
\label{tab:main_federated_results}
\end{table*}

Table~\ref{tab:main_federated_results} reports the collaborative-method comparison. Table~\ref{tab:phase1-main-results} separates the residual parameterization from the alignment loss. For BlogText, author accuracy, macro-F1, and verification AUC/EER are all computed in the frozen ASCE space. Verification AUC/EER uses the same-author versus the different-author generated-continuation pairs.

Table~\ref{tab:phase1-main-external-verification} reports
ASCE-independent generated-to-source cosine verification.
StyleDist. denotes StyleDistance embeddings, MiniLM denotes
all-MiniLM-L6-v2 sentence embeddings, and Stylom. denotes classical
stylometric features (character/word $n$-grams, function-word
frequencies, and punctuation/structure statistics).
Same-author generated/source pairs are positives and same-topic,
different-author pairs are negatives; AUC and EER denote ROC area
and equal error rate, respectively. Appendix~\ref{subsec:experiment_asce_leakage_free_audit} reports the expanded BlogText ASCE results, including source-anchored and generated-text views, and Appendix~\ref{subsec:experiment_significance} reports paired tests.

\begin{table}[h!]
\centering
\footnotesize
\setlength{\tabcolsep}{2.2pt}
\renewcommand{\arraystretch}{1.07}
\begin{tabular}{@{}lcccc@{}}
\toprule
Method & \makecell{StyleDist\\AUC $\uparrow$} & \makecell{StyleDist\\EER $\downarrow$} & \makecell{MiniLM\\AUC $\uparrow$} & \makecell{Stylom.\\AUC $\uparrow$} \\
\midrule
FedProx & 0.724 & 0.334 & 0.631 & 0.651 \\
Pooled PEFT & 0.736 & 0.327 & 0.650 & 0.660 \\
Ditto & 0.748 & 0.318 & 0.650 & 0.683 \\
FedDPA\textsuperscript{\dag} & 0.747 & 0.318 & 0.659 & 0.692 \\
Local-only + ASCE & \underline{0.768} & \underline{0.301} & \textbf{0.685} & \underline{0.708} \\
FAVoR & \textbf{0.771} & \textbf{0.295} & \underline{0.675} & \textbf{0.714} \\
\bottomrule
\end{tabular}
\caption{Main BlogText results with external authorship verification.}
\label{tab:phase1-main-external-verification}
\end{table}

Across both datasets, FAVoR shows the strongest collaborative author-style retention, with small, mixed changes in continuation utility. On BlogText, slight BERTScore decreases against Ditto/FedProx/FedDPA are outweighed by larger retention gains, and FAVoR also improves over pooled PEFT in both author accuracy and BERTScore. Table~\ref{tab:phase1-main-external-verification} confirms FAVoR's author-style-retention advantage over the collaborative baselines under ASCE-independent verifiers, while local-only remains a strong non-collaborative reference. The Mythos-Reddit breakdown appears in Appendix~\ref{subsec:experiment_mythos_external}.

A method-blinded reference-free Qwen judge found no significant generic-quality difference between FAVoR and any collaborative baseline after Holm correction. In a separate pairwise author-voice test, the judge preferred FAVoR in $61.2\%$ of non-ties (author-clustered $p=0.022$), supporting author-style fidelity rather than generic utility superiority; Appendix~\ref{subsec:experiment_llm_judges} gives the protocol and full results.

A small human 2AFC audit provides a complementary perceptual check: FAVoR continuations were matched to the intended author more often than Ditto continuations (91.7\% vs. 68.3\% vote-level accuracy).
We use this only as supporting evidence that the retained author signal is human-perceivable, not as a standalone preference or authenticity evaluation. Details and qualitative examples appear in Appendices~\ref{subsec:human_2afc_audit} and~\ref{subsec:experiment_qualitative}.

\begin{table}[h!]
\centering
\footnotesize
\setlength{\tabcolsep}{2.6pt}
\renewcommand{\arraystretch}{1.07}
\begin{tabular}{@{}lcccc@{}}
\toprule
Variant & \makecell{Author\\Acc. $\uparrow$} & \makecell{Macro-\\F1 $\uparrow$} &
\makecell{BERTScore\\$\uparrow$} & \makecell{StyleDist.\\AUC $\uparrow$} \\
\midrule
Global ASCE & 0.161 & 0.140 & \textbf{0.818} & 0.728 \\
Ditto + ASCE & 0.318 & 0.286 & 0.816 & 0.746 \\
FAVoR w/o ASCE & 0.377 & 0.331 & 0.814 & \underline{0.767} \\
FAVoR & \textbf{0.403} & \textbf{0.365} & 0.815 & \textbf{0.771} \\
\bottomrule
\end{tabular}
\caption{Core BlogText component ablations with StyleDistance gen$\rightarrow$source verification.}
\label{tab:phase1-main-results}
\end{table}

\paragraph{Ablations.}
Table~\ref{tab:phase1-main-results} reports FAVoR component ablations and a targeted Ditto+ASCE retrofit.
The Ditto+ASCE row keeps Ditto's personalized proximal branch but adds the same local ASCE alignment loss, testing whether the gains can be explained by simply attaching ASCE to a personalized-FL baseline.
The FAVoR w/o ASCE row removes the alignment loss while preserving the shared--private residual decomposition.
Thus, the table separates the contribution of style alignment from the contribution of the residual aggregation constraint.

\begin{figure}[h!]
    \centering
    \includegraphics[width=0.98\linewidth]{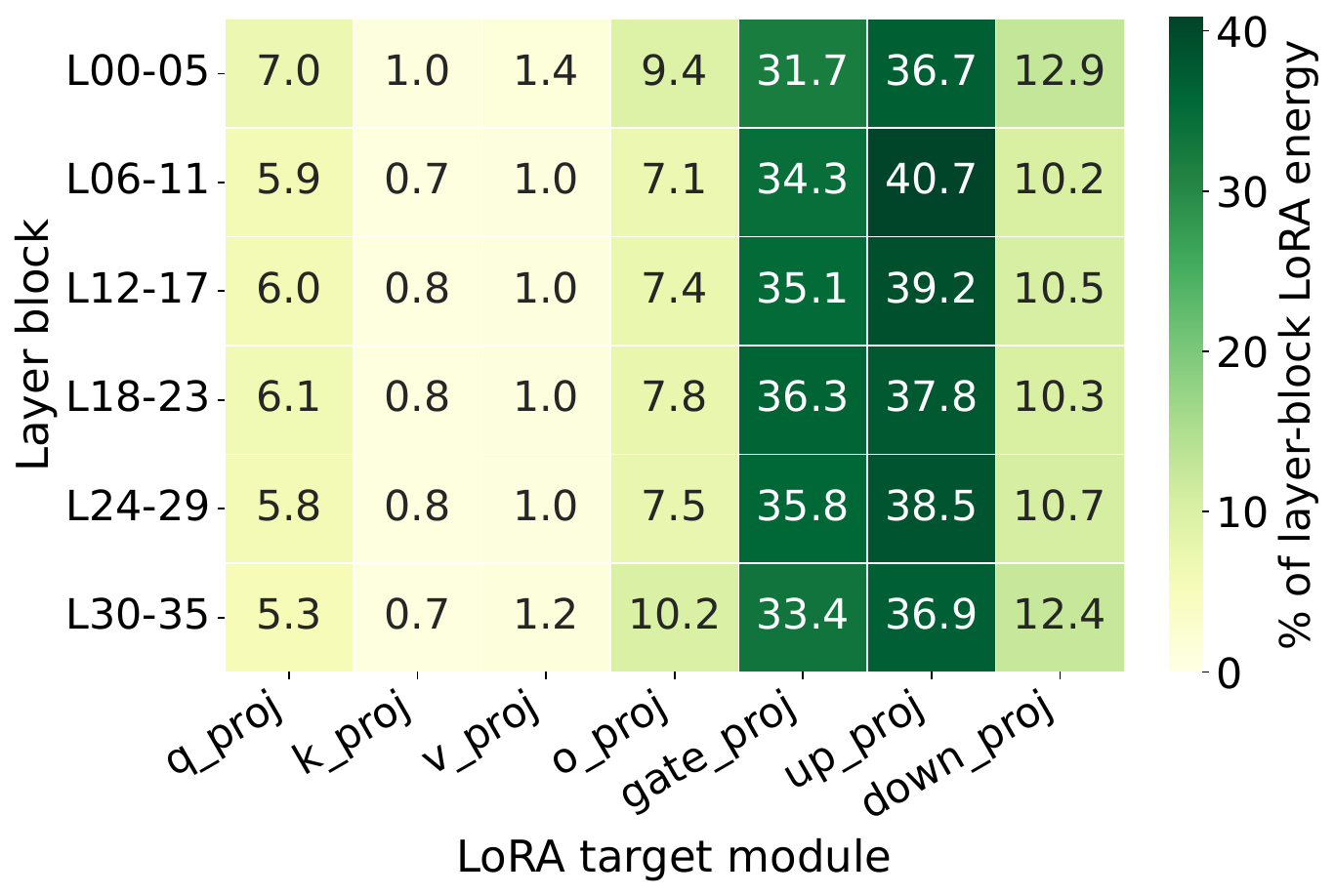}
    \caption{Adapter-space diagnostic for FAVoR on BlogText. Cells show normalized LoRA update energy by layer block and module, averaged over 50 personalized adapters.}
    \label{fig:favor_lora_allocation}
\end{figure}

The two components are separable but not interchangeable. Attaching the same ASCE alignment loss to a personalized-FL baseline (Ditto+ASCE) improves over the shared-only Global ASCE control but recovers only a fraction of FAVoR's author-accuracy gain, whereas FAVoR without the alignment loss already attains most of it. The shared--private aggregation protocol, not the alignment loss alone, accounts for the bulk of author-style retention. Alignment provides a smaller, metric-dependent change on top.
The main-text table reports seed mean. Appendix~\ref{subsec:ablation_significance_audit} reports the mean$\pm$standard-deviation version, paired ablation tests, external-verifier tests, and configuration checks.
These tests support the author-style gains while showing that utility differences are small and mixed.
Source-anchored ASCE mean$\pm$standard-deviation diagnostics are reported in Appendix~\ref{subsec:experiment_asce_leakage_free_audit}, and the 100-author stress test is reported in Appendix~\ref{subsec:experiment_k100_scale}.
An independently trained ASCE-B post-hoc evaluation also preserves the ordering: FAVoR reaches source-prototype accuracy/AUC of $0.403/0.771$, versus $0.214/0.723$ for Ditto. Controlled text-volume, author-imbalance, and ASCE evaluator-scale checks retain FAVoR's relative author-retention advantage within the tested ranges; Appendices~\ref{subsec:data_scale_imbalance}, \ref{subsec:asce_b_audit}, and~\ref{subsec:asce_scale_transfer} report these diagnostics.
\paragraph{Adapter-space diagnostic.}
Figure~\ref{fig:favor_lora_allocation} shows that personalized FAVoR adapters concentrate LoRA update energy in multilayer perceptron (MLP) projections rather than distributing it uniformly across target modules, consistent with the private residual pack storing structured corrections on top of the shared endpoint (details in Appendix~\ref{subsec:adapter_space_diagnostic}).

\subsection{Cold-Start Transfer to Unseen Authors}
\label{subsec:experiment_cold_start}

We use cold-start transfer as a supplementary test of whether the federated shared adapter provides a reusable writing initialization.
The shared adapter is trained on 40 federated authors and transferred to 10 held-out authors.  Each held-out author then trains a lightweight private residual from only $K$ local examples and is evaluated on disjoint continuations.
To avoid author-label evaluator leakage, cold-start author-separability metrics use a separate ASCE-style encoder trained without the 10 held-out authors and classify generations using nearest-prototype matching to held-out-author prototypes built only from the same $K$ adaptation examples.
Table~\ref{tab:cold-start-main} reports the compact $K=8$ no-ASCE private-adaptation variant, isolating the effect of the learned shared initialization. ASCE variants are reported in Appendix~\ref{subsec:experiment_cold_start_details}.

\begin{table}[t]
\centering
\footnotesize
\setlength{\tabcolsep}{3.2pt}
\renewcommand{\arraystretch}{1.06}
\begin{tabularx}{\columnwidth}{@{}>{\raggedright\arraybackslash}Xccc@{}}
\toprule
Method & \makecell{Sep.\\Acc. $\uparrow$} & \makecell{Macro-\\F1 $\uparrow$} &
\makecell{Target\\Sim. $\uparrow$} \\
\midrule
Shared only & 0.150 & 0.166 & 0.191 \\
Random shared init + private & 0.308 & 0.278 & 0.201 \\
Local-only from base & 0.425 & 0.428 & 0.259 \\
FAVoR shared init + private & \textbf{0.641} & \textbf{0.629} & \textbf{0.363} \\
\bottomrule
\end{tabularx}
\caption{$K=8$ cold-start author-separability for unseen authors. FAVoR uses no-ASCE private adaptation.}
\label{tab:cold-start-main}
\end{table}

At $K=8$, FAVoR improves over local-only adaptation from the base model in author accuracy, macro-F1, and target similarity.
It also outperforms the random-shared-initialization and shared-only controls, indicating that the gain depends on both a learned federated initialization and private adaptation.
A nested no-ASCE rerun over $K\in\{4,8,12,16,24\}$ confirms that the shared-initialization advantage is clearest at $K=8$: with fixed 32-text evaluation prototypes, FAVoR exceeds local-only accuracy by $0.119$ (95\% CI $[0.050,0.203]$), while neither method is reliably superior at $K=16$--$24$.
We therefore interpret federation as most useful in the tested low-resource regime, not as establishing a universal crossover threshold. The full $K=8$ and $K=16$ evaluation and the nested $K$-sweep are reported in Appendix~\ref{subsec:experiment_cold_start_details}.

\section{Conclusion}
\label{sec:conclusion}

We studied author-style homogenization as a utility-preserving failure mode in federated personalized continuation and introduced an evaluation protocol that reveals it beyond standard utility metrics such as BERTScore and MAUVE.
FAVoR instantiates an author-style residual mechanism on a shared-private PEFT design: shared updates are aggregated, while local-style residuals remain outside the aggregation.
Empirically, standard federated PEFT can erase cross-author style even when utility metrics remain acceptable. FAVoR mitigates this mainly through the residual aggregation boundary rather than the style-alignment loss alone.
Future work should extend the analysis to larger author populations, broader writing domains, and stronger privacy mechanisms for author-specific adaptation.

\section*{Limitations}
\label{sec:limitations}

Our evaluation treats author style as an operational construct measured through held-out continuation, learned authorship spaces, and external gen$\rightarrow$source verification.
These metrics capture author-specific stylistic signals, but they do not exhaust social, pragmatic, or reader-perceived notions of authorial voice; human judgments of perceived authenticity are outside the scope of this work.
The BlogText ASCE encoder is trained centrally before FL on authors disjoint from the 50-author FL roster, rather than learned through a federated or differentially private procedure.
This removes main-roster author overlap for both the local style-alignment target and the ASCE-space evaluator, but learned style spaces can still encode topic, genre, demographic artifacts, and other corpus-specific regularities.
We therefore pair ASCE-space metrics with ASCE-independent external verifiers and report the ASCE training split explicitly.

Our privacy claim is limited to data locality during the FL procedure.
We do not provide formal differential privacy, secure aggregation, or resistance to gradient- or adapter-inversion attacks, and we do not upload or release private residual packs.
The experiments are also limited to English public writing domains and to a single backbone family.
The 7B result is a single-seed sensitivity probe and changes both model scale and instruction-tuning status, from 3B-Instruct to 7B-base, so it should be read as robustness evidence rather than a clean scaling ablation.

\section*{Ethical Considerations}
\label{sec:ethical_considerations}

\paragraph{Consent, licensing, and data use.}
We use public research corpora and public evaluation artifacts only for non-commercial research.
The main benchmark is a filtered split of the Blog Authorship Corpus distributed through Tatman's Kaggle redistribution. The external validation uses the Reddit/WritingPrompts portion of Mythos-Reddit \citep{schler-etal-2006-blogging,tatman-blog-authorship-corpus-kaggle,ashok-kumar-etal-2025-whose}.
Public availability should not be treated as blanket consent for author-style modelling or imitation.
Because author identity and author-specific style are central to the task, the underlying public corpora may contain names, handles, author-linkable details, or offensive content.
We do not conduct a comprehensive audit of personally identifiable information (PII)/offensive-content audit or attempt full de-identification of the source corpora.
Instead, for reporting and release, we report only aggregate metrics, anonymize handles and signatures in qualitative examples, do not infer protected attributes, and do not publicly redistribute raw author texts, author-reference sets, author targets, user handles, or author-linkable generated continuations.
Users of any released code or scripts must obtain the underlying datasets from their original sources and comply with the corresponding dataset licenses, platform terms, and deletion requests where applicable.

\paragraph{Human annotation.}
The human 2AFC audit involved three annotators who evaluated anonymized author-profile excerpts and candidate continuations.
Annotators provided informed consent and were compensated for their time.
Annotators did not receive author handles, demographic metadata, source URLs, or author identities, and we collected only their 2AFC choices, confidence ratings, and optional notes.
We report only aggregate results.
The audit was used as a small perceptual attribution check rather than a preference, authenticity, or quality evaluation.
The audit fell under ``quality assurance and evaluation activities'' according to our institutional human ethics policy and therefore did not require ethics approval.

\paragraph{Model, evaluator, and release policy.}
Experiments use Qwen2.5-3B-Instruct as the backbone and public evaluator artifacts, including StyleDistance, MiniLM, BERTScore, MAUVE, and stylometric features \citep{qwen25-3b-instruct,patel-etal-2025-styledistance,sentence-transformers-minilm-l6-v2,zhang-etal-2020-bertscore,JMLR:v24:23-0023}.
Code, configuration files, preprocessing/evaluation scripts, and aggregate evaluation outputs are available at \url{https://github.com/vermouth369/favor-author-style}.
We do not release the base-model weights, raw or author-linkable generated texts, author targets, private residual packs, or author-specific adapter checkpoints.
If trained shared adapters are released, they will be distributed only when compatible with the base-model license and the relevant data-source terms.
Because author-style adaptation can support deceptive imitation, deployment should require author consent, disclosure to affected users, access controls, and safeguards against generating text under another person's identity without permission.

\bibliography{custom}

\clearpage
\appendix

\section{FAVoR Training Details}
\label{app:favor_training_details}

\subsection{Complete Training Pseudocode}
\label{subsec:favor_full_algorithm}

Algorithm~\ref{alg:favor_full} expands the compact round summary in Algorithm~\ref{alg:favor_main}.
It makes explicit which quantities remain client-local and which updates are uploaded for aggregation.

\begin{algorithm}[h!]
\caption{FAVoR: Federated Shared-Private Training with ASCE Alignment}
\label{alg:favor_full}
\footnotesize
\begin{algorithmic}[1]
\Require Frozen base model $\theta$; fixed ASCE encoder $E_{\phi}$; initial shared adapter $A^{(g)}_0$
\Require Local private residual-pack store; local corpora $\{\mathcal{D}_i^{\mathrm{train}}\}_{i=1}^N$; local reference sets $\{\mathcal{D}_i^{\mathrm{ref}}\}_{i=1}^N$
\For{$t = 0,1,\dots,T-1$}
    \State Server samples clients $\mathcal{S}_t$
    \State Server broadcasts shared adapter $A^{(g)}_t$ to clients in $\mathcal{S}_t$
    \ForAll{clients $i \in \mathcal{S}_t$ in parallel}
        \State Construct local author target $q_i$ from $\mathcal{D}_i^{\mathrm{ref}}$ using Eq.~\eqref{eq:author_target}
        \State Initialize local shared model $\theta \oplus A^{(g)}_t$
        \State Update the local shared copy using $\mathcal{L}^{\mathrm{shared}}_i$ in Eq.~\eqref{eq:shared_objective}, yielding $A_{i,t}^{(g)}$
        \State Compute shared-adapter delta $\Delta A_{i,t}^{(g)}$ using Eq.~\eqref{eq:shared_delta}
        \State Initialize the private adapter from $A_{i,t}^{(g)}$; then, holding $A_{i,t}^{(g)}$ fixed, update the private adapter using $\mathcal{L}^{\mathrm{private}}_i$ in Eq.~\eqref{eq:private_objective}
        \State Store the local shared endpoint $A_{i,t}^{(g)}$ and updated private residual pack $\Delta A_{i,t}^{(p)}$ using Eq.~\eqref{eq:residual_pack}
        \State Upload only $\Delta A_{i,t}^{(g)}$ to the server
    \EndFor
    \State Server aggregates shared deltas using Eq.~\eqref{eq:shared_aggregation_delta}
    \State Server updates shared adapter using Eq.~\eqref{eq:shared_adapter_update}
\EndFor
\State \Return Final shared adapter $A^{(g)}_T$ and retained client states $\{A_{i,\tau_i}^{(g)}, \Delta A_{i,\tau_i}^{(p)}\}_{i=1}^N$
\end{algorithmic}
\end{algorithm}

\section{Additional Experiments and Diagnostics}
\label{app:additional_experiments}

\subsection{Experimental Protocol and Metric Definitions}
\label{subsec:appendix_protocol_definitions}

The cross-dataset comparison tests whether FAVoR's author-style-retention advantage over the collaborative baselines replicates beyond BlogText.
FedAvg averages sampled-client LoRA deltas using uniform client weighting.
FedProx adds a proximal penalty toward the global adapter.
Our pFedMe-style baseline optimizes a personalized adapter around a client meta/global anchor and uploads a meta-update.
Ditto maintains a global adapter and a locally personalized adapter regularized toward it; only the global update is uploaded.
FedDPA mixes global and local adapters through a learned gate; reported generations use the client mean validation gate, with dynamic gating audited separately in Appendix~\ref{subsec:feddpa_gate_sensitivity}.

Topic control restricts comparisons rather than making representations topic-free.
Author-within-topic effect sizes are computed within metadata-topic strata, while strict source verification and generated-to-source verification use same-author positives and same-topic, different-author negatives.
Probe accuracy and macro-F1 are global author-classification metrics and are not topic-controlled.
Table~\ref{tab:main_federated_results} accuracy and macro-F1 use nearest source-train prototypes, and its AUC/EER use generated-to-generated verification; Table~\ref{tab:phase1-main-external-verification} uses generated-to-source verification.
Higher AUC and lower EER are better.

Each nearest-author prototype is the normalized mean of the permitted normalized source embeddings, and a generation is assigned to the prototype with highest cosine similarity.
The main evaluation uses source-train texts; cold-start evaluation uses only the same $K$ adaptation examples.
This prototype assignment rule is not used for Table~\ref{tab:signature-space-main} or Table~\ref{tab:phase1-main-external-verification}.
The no-ASCE ``Local-only from base'' cold-start baseline trains one LoRA adapter per held-out author from the frozen Qwen2.5-3B-Instruct backbone using the same $K$ adaptation examples and budget, without federated initialization or cross-client aggregation; evaluation continuations are disjoint from the adaptation examples.

BERTScore and SemSim are paired reference-based metrics comparing each generation with its matched human continuation; SemSim is the cosine similarity between all-MiniLM-L6-v2 embeddings.
MAUVE instead compares aggregate generated and human continuation distributions.
The main table reports BERTScore and SemSim, while the appendix tables additionally report MAUVE.

\subsection{Qualitative Error Analysis}
\label{subsec:experiment_qualitative}

Table~\ref{tab:qualitative-appendix} reports a matched qualitative comparison referenced from the main text.
Examples are sampled from the common post-filtering held-out continuation set using a fixed seed (4302026), after matching seed, author, and prompt across all methods in the table.
The common intersection contains 190 examples.
We use a fixed method order and show shortened raw continuations side by side; no additional training, decoding, or post-hoc rationales are added.
The public table uses anonymous example indices and shortened excerpts.
The examples are intended to make the measured homogenization failure mode legible at the text level, while the aggregate conclusions are based on the quantitative metrics.

\begin{table*}[t]
\centering
\scriptsize
\setlength{\tabcolsep}{3pt}
\renewcommand{\arraystretch}{1.03}
\begin{tabularx}{\textwidth}{@{}p{0.14\textwidth}p{0.15\textwidth}>{\raggedright\arraybackslash}X@{}}
\toprule
Example & Method & Shortened raw continuation \\
\midrule
\multicolumn{3}{@{}p{\textwidth}@{}}{\textbf{1}. Prefix cue: ``Now I can post about other things. Yes I'm prepared for the hurricane. Had to go back to Lowes today to get some things. My gas \ldots'' Gold cue: ``They don't want the kids to get stuck at school and the road flood and then not \ldots''} \\
 & FedAvg & ``We will be staying home on Monday, Tuesday, Wednesday, and Thursday. Friday is a day off. So we should be fine. \ldots'' \\
 & Pooled PEFT & ``We will see what happens. I think my mom is going to stay with us until Saturday. She has a lot \ldots'' \\
 & FedProx & ``We will be staying home on Monday, Tuesday, Wednesday, and Thursday. Friday is a work day but they will probably close \ldots'' \\
 & Ditto & ``We will see what happens. I am glad you guys made it through ok. It's nice to hear from you again. \ldots'' \\
 & FAVoR & ``We will see what happens. I am glad you guys had a good time at the beach. It is nice to \ldots'' \\
 & Local-only + ASCE & ``We will see what happens. I am going to try to keep up with posting but I know how hard it \ldots'' \\
\midrule
\multicolumn{3}{@{}p{\textwidth}@{}}{\textbf{2}. Prefix cue: ``When I started high school I didn't really know anybody. They had just build HCHS near my house and I was part of the first freshman \ldots'' Gold cue: ``met Brent, and through Brent I met Mike and Brad. And through Brad I met Hans. Brent lived \ldots''} \\
 & FedAvg & ``don't remember what year it was exactly but I think it was either 1986 or 1879. Anyway, when I got there I \ldots'' \\
 & Pooled PEFT & ``don't remember how many people were in the freshman class, but it wasn't very big. We all lived within a mile or \ldots'' \\
 & FedProx & ``don't remember what year it was, but I think it was either 1986 or 1879. Anyway, I met a lot of people \ldots'' \\
 & Ditto & ``think it's a good school. It has a lot of clubs and activities. There are also some pretty nice teachers. I'm not \ldots'' \\
 & FAVoR & ``think it's still called that now. It was a small school with about 100 kids in each grade. There were no sports \ldots'' \\
 & Local-only + ASCE & ``don't remember what year it was but I think it was either '82 or '83. Anyway, when I got there I met \ldots'' \\
\midrule
\multicolumn{3}{@{}p{\textwidth}@{}}{\textbf{3}. Prefix cue: ``omg last night was cotillion ball...omg i fainted!!!lol when me ashleigh brett and sammie were leavin i made it down one step and fainted down two..lol \ldots'' Gold cue: ``ANG...(ONLY THE OTHER WOR) UVE CHANGED and he had this HUGE grin on his face..anyways one time when \ldots''} \\
 & FedAvg & ``AMN IT! lol so anyway we got home and i took a nap...then woke up at 10:30 and went to the mall \ldots'' \\
 & Pooled PEFT & ``ONT SAY ANYTHING TO HER! lol so thats what happened..lol but anyways its over now..i have a test tomorrow on the stuff \ldots'' \\
 & FedProx & ``AMN IT! lol so anyway we got home and i took a nap...i woke up at 10:30 and then went to the \ldots'' \\
 & Ditto & ``IDN'T YOU TELL ME THAT LAST YEAR? lol thats so funny...he said that like its a big deal...i mean come on, we've \ldots'' \\
 & FAVoR & ``IDN'T YOU KNOW THAT? lol thats so funny..anyways we got out of the limo and walked in..i think that's how they do \ldots'' \\
 & Local-only + ASCE & ``IDNT YOU TELL ME THAT LAST WEEK? lol thats what he said..i think he meant that as a compliment but still..lol anyways \ldots'' \\
\bottomrule
\end{tabularx}
\caption{Matched qualitative comparison over anonymized held-out continuation examples. Entries are shortened raw generated continuations from the same seed-author-prompt instances across all listed methods. The table shows three examples from the fixed-seed sample.}
\label{tab:qualitative-appendix}
\end{table*}

\subsection{Adapter-Space Diagnostic}
\label{subsec:adapter_space_diagnostic}

Figure~\ref{fig:favor_lora_allocation} provides an auxiliary parameter-space diagnostic for the shared--private decomposition.
Using 50 personalized adapters from one representative BlogText run, we define update energy as the squared Frobenius norm of the effective LoRA update $\Delta W_\ell=(\alpha/r)V_\ell U_\ell$.
Energies are aggregated by six-layer block and target module and summarized as normalized percentage distributions across the seven target modules; each heatmap row sums to approximately 100\%, up to rounding.
The pattern is not uniform: most updated energy is concentrated in MLP projections, especially \texttt{gate\_proj} and \texttt{up\_proj}, while \texttt{k\_proj} and \texttt{v\_proj} receive comparatively little energy.
This is a supportive parameter-space diagnostic rather than causal mechanism evidence; the main mechanism evidence comes from the component ablations in Table~\ref{tab:phase1-main-results}.

\subsection{Prompting and Retrieval Baselines}
\label{subsec:experiment_prompting_baselines}

Table~\ref{tab:prompting_retrieval_baselines} compares FAVoR with non-training personalization baselines inspired by Mythos-style author artifacts.
These baselines test whether raw author history, retrieval-augmented generation (RAG), or distilled author sheets can recover author style without federated optimization.
They serve as supplementary non-training controls; the primary collaborative-learning comparison is reported in Tables~\ref{tab:main_federated_results} and~\ref{tab:phase1-main-external-verification}.
Because the prompting baselines were run on a fixed prompting-aligned subset, Table~\ref{tab:prompting_retrieval_baselines} reports a source-anchored author-disjoint ASCE rescore of those generated outputs.
The prototype and gen$\rightarrow$source columns follow the source-anchored view in Table~\ref{tab:asce-leakage-free-gensource}, but are computed only on this imported prompting-baseline subset; they are not the generated-text Author Acc. values reported in Table~\ref{tab:main_federated_results}.

\begin{table*}[t]
\centering
\footnotesize
\renewcommand{\arraystretch}{1.06}

\textit{Panel A: Configuration and author-style metrics}\par\vspace{2pt}

\setlength{\tabcolsep}{1.4pt}
\begin{tabularx}{\textwidth}{
@{}>{\raggedright\arraybackslash}X
cccccccccc@{}}
\toprule
Method &
FL &
Train &
\makecell{Raw\\Hist.} &
RAG &
\makecell{Distilled\\Artifact} &
\makecell{Proto.\\Acc. $\uparrow$} &
\makecell{Proto.\\F1 $\uparrow$} &
\makecell{Gen$\rightarrow$Src\\AUC $\uparrow$} &
\makecell{EER\\$\downarrow$} &
\makecell{Silh.\\$\uparrow$} \\
\midrule
FAVoR &
\yesmark & \yesmark & \nomark & \nomark & \nomark &
\textbf{0.288} & \textbf{0.208} & \textbf{0.760} &
\textbf{0.319} & \textbf{-0.045} \\

Mythos Avg-Author &
\nomark & \nomark & \nomark & \nomark & \nomark &
0.093 & 0.047 & 0.627 & 0.394 & -0.184 \\

Mythos RAG-Only &
\nomark & \nomark & \yesmark & \yesmark & \nomark &
0.131 & 0.095 & 0.713 & 0.372 & -0.198 \\

Mythos Sheet + RAG &
\nomark & \nomark & \yesmark & \yesmark & \yesmark &
0.176 & 0.126 & 0.689 & 0.389 & -0.145 \\

Mythos Sheet (no persona) + RAG &
\nomark & \nomark & \yesmark & \yesmark & \yesmark &
0.130 & 0.088 & 0.698 & 0.365 & -0.158 \\

Mythos Sheet Only &
\nomark & \nomark & \nomark & \nomark & \yesmark &
0.150 & 0.103 & 0.665 & 0.378 & -0.143 \\
\bottomrule
\end{tabularx}

\vspace{5pt}
\textit{Panel B: Continuation-quality metrics}\par\vspace{2pt}

\setlength{\tabcolsep}{6pt}
\begin{tabularx}{\textwidth}{
@{}>{\raggedright\arraybackslash}Xccc@{}}
\toprule
Method &
BERTScore $\uparrow$ &
SemSim $\uparrow$ &
MAUVE $\uparrow$ \\
\midrule
FAVoR                    & 0.815 & 0.385 & \textbf{0.313} \\
Mythos Avg-Author        & 0.814 & 0.395 & 0.013 \\
Mythos RAG-Only          & \textbf{0.819} & 0.405 & 0.025 \\
Mythos Sheet + RAG       & 0.815 & \textbf{0.415} & 0.093 \\
Mythos Sheet (no persona) + RAG & 0.816 & 0.411 & 0.030 \\
Mythos Sheet Only        & 0.812 & 0.396 & 0.061 \\
\bottomrule
\end{tabularx}

\caption{Prompting and retrieval baselines on BlogText.
Author-style columns use the author-disjoint ASCE source-anchored
rescore on the shared prompting-aligned subset for this supplemental
comparison. Prototype accuracy/F1 classifies generated continuations
by the nearest source-train author prototype, and
Gen$\rightarrow$Src AUC/EER uses same-author versus different-author
generated-to-source pairs. Utility columns report the original
generation-quality metrics for the same comparison.}
\label{tab:prompting_retrieval_baselines}
\end{table*}

\subsection{Source-Space Controls}
\label{subsec:experiment_source_controls}

Table~\ref{tab:signature-space-native} reports the native-dimensional version of the source-space diagnostic introduced in Section~\ref{subsec:experiment_signature_space}.
Together with Table~\ref{tab:signature-space-main}, it checks whether the internal ASCE space is competitive with external authorship/style encoders and whether the author signal remains visible under topic-controlled verification.
This diagnostic is a held-out comparison of frozen RoBERTa, StyleDistance, LUAR, STAR, and ASCE representations.
For each representation, a class-balanced multinomial logistic-regression probe is fitted on source-train embeddings, its $C$ is selected by source-validation macro-F1, and accuracy and macro-F1 are reported on source-test.
AUC and EER are separate cosine-verification metrics using same-author positives and same-topic, different-author negatives.
Author-within-topic $\omega^2$ is the variance explained by author within metadata-topic strata, whereas topic $\omega^2$ is the variance explained by topic; the A/T ratio is their ratio.
The topic--author origin-fit slope in Figure~\ref{fig:signature_space_discriminability} is the origin-constrained slope relating the coordinate-level topic and author effect sizes.

\begin{table*}[t]
\centering
\footnotesize
\setlength{\tabcolsep}{1.8pt}
\renewcommand{\arraystretch}{1.08}

\begin{tabularx}{\textwidth}{
@{}>{\raggedright\arraybackslash}Xrccccccc@{}}
\toprule
Representation space &
Dim. &
\makecell{Author\\Acc. $\uparrow$} &
\makecell{Macro-\\F1 $\uparrow$} &
\makecell{Author-in-topic\\$\omega^2$ $\uparrow$} &
\makecell{Topic\\$\omega^2$ $\downarrow$} &
\makecell{A/T\\ratio $\uparrow$} &
\makecell{Strict\\AUC $\uparrow$} &
\makecell{EER\\$\downarrow$} \\
\midrule
RoBERTa-base
& 768 & 0.635 & 0.677 & 0.1135 & 0.0557 & 1.95 & 0.629 & 0.421 \\

StyleDistance
& 768 & 0.628 & 0.660 & \textbf{0.2357} & 0.0577
& \textbf{4.21} & 0.716 & 0.344 \\

LUAR-MUD doc1
& 512 & 0.633 & 0.677 & 0.1257 & \textbf{0.0364}
& 3.52 & 0.797 & 0.277 \\

STAR
& 1024 & \underline{0.648} & \textbf{0.715}
& 0.1886 & \underline{0.0478} & \underline{4.16}
& \underline{0.856} & \underline{0.229} \\

ASCE Author-Style Signature Space
& 256 & \textbf{0.651} & \underline{0.692}
& \underline{0.1916} & 0.0499 & 3.64
& \textbf{0.863} & \textbf{0.211} \\
\bottomrule
\end{tabularx}

\caption{Native-dimensional source-space diagnostics.}
\label{tab:signature-space-native}
\end{table*}

\subsection{7B Base-Model Sensitivity}
\label{subsec:experiment_7b_base_sensitivity}

Table~\ref{tab:phase1-7b-base-sensitivity} reports a single-seed backbone-sensitivity probe on the same 50-author medium non-IID BlogText split.
This run replaces the main 3B instruction-tuned backbone with a Qwen2.5-7B base model \citep{qwen25-3b-instruct} and keeps the held-out continuation protocol, LoRA recipe, decoding, ASCE evaluator, and full $K=50$ metric pass fixed.
Because this changes both parameter scale and instruction-tuning status, and because it uses only one seed, we treat it as robustness evidence rather than a main multi-seed result.

\begin{table*}[t]
\centering
\scriptsize
\setlength{\tabcolsep}{3pt}
\renewcommand{\arraystretch}{1.06}
\resizebox{\textwidth}{!}{%
\begin{tabular}{lccccccccc}
\toprule
Method & \makecell{Author\\Acc. $\uparrow$} & \makecell{Macro-\\F1 $\uparrow$} & \makecell{Verif.\\AUC $\uparrow$} & \makecell{EER\\$\downarrow$} & \makecell{Silhouette\\$\uparrow$} & \makecell{Asst.\\Score $\downarrow$} & \makecell{BERTScore\\$\uparrow$} & \makecell{SemSim\\$\uparrow$} & \makecell{MAUVE\\$\uparrow$} \\
\midrule
Base only & 0.207 & 0.191 & 0.695 & 0.353 & -0.141 & 0.865 & 0.815 & 0.353 & 0.130 \\
FedAvg & 0.219 & 0.170 & 0.716 & 0.343 & -0.136 & 0.230 & 0.811 & 0.361 & 0.263 \\
FedProx & 0.214 & 0.177 & 0.698 & 0.365 & -0.149 & \textbf{0.183} & 0.808 & 0.352 & 0.327 \\
pFedMe & 0.293 & 0.245 & 0.807 & 0.263 & -0.053 & 0.625 & \textbf{0.816} & 0.365 & 0.327 \\
Ditto & 0.368 & 0.328 & 0.836 & 0.243 & -0.033 & 0.424 & 0.813 & \textbf{0.381} & \textbf{0.417} \\
FAVoR & \textbf{0.528} & \textbf{0.487} & \textbf{0.899} & \textbf{0.185} & \textbf{0.134} & 0.251 & 0.810 & 0.378 & 0.334 \\
\bottomrule
\end{tabular}%
}
\caption{Single-seed Qwen2.5-7B base-model sensitivity probe on the full 50-author BlogText held-out continuation split. Bold marks the best listed value; base-only is a non-collaborative reference.}
\label{tab:phase1-7b-base-sensitivity}
\end{table*}

The ordering matches the main finding.
FAVoR is the strongest collaborative method on every author-style retention column.
Relative to FedAvg, FAVoR improves author accuracy by 0.310 and verification AUC by 0.183, while retaining all generated continuations after filtering.

\subsection{100-Author Scale Stress Test}
\label{subsec:experiment_k100_scale}

To check whether the main 50-author results depend on a small roster, we ran an additional BlogText stress test with 100 authors over three seeds.
The roster preserves the original 50 authors and adds 50 authors using the same style-cluster balancing procedure, with the same per-author token caps as the main data construction.
This experiment is meant as scaling evidence rather than a new primary significance table: it uses $(R, C,\bar{p})=(20,20,4.0)$ rounds, clients per round, and expected participations per author, compared with $(6,12,1.44)$ in the main 50-author BlogText runs.

Table~\ref{tab:k100-scale-stress} shows that the main ordering largely persists at the larger author scale.
Shared-only FedAvg has low author separability, negative silhouette, and high verification error, indicating strong author-style homogenization.
FAVoR substantially mitigates this collapse, recovering roughly 70\% of the FedAvg-to-local-only gap in separability accuracy, 74\% in macro-F1, 81\% in verification AUC, and 72\% in silhouette.
The result should still be read conservatively: FAVoR is close to Ditto on verification AUC/EER and stronger on closed-set author separability, while Ditto obtains higher MAUVE.
Local-only remains the non-collaborative style retention reference.

\begin{table*}[t]
\centering
\footnotesize
\setlength{\tabcolsep}{1.8pt}
\renewcommand{\arraystretch}{1.14}

\begin{tabularx}{\textwidth}{
@{}>{\raggedright\arraybackslash}Xcccccccc@{}}
\toprule
Method &
\makecell{Sep.\\Acc. $\uparrow$} &
\makecell{Macro-\\F1 $\uparrow$} &
\makecell{Verif.\\AUC $\uparrow$} &
\makecell{EER\\$\downarrow$} &
\makecell{Centroid\\Dist. $\uparrow$} &
\makecell{Silhouette\\$\uparrow$} &
\makecell{SemSim\\$\uparrow$} &
\makecell{MAUVE\\$\uparrow$} \\
\midrule
FedAvg shared &
\meanstd{0.177}{0.013} &
\meanstd{0.150}{0.013} &
\meanstd{0.729}{0.002} &
\meanstd{0.331}{0.002} &
\meanstd{0.677}{0.007} &
\meanstd{-0.176}{0.010} &
\meanstd{0.385}{0.006} &
\meanstd{0.272}{0.040} \\

Ditto &
\meanstd{0.386}{0.022} &
\meanstd{0.342}{0.017} &
\meanstd{0.883}{0.012} &
\meanstd{0.193}{0.008} &
\meanstd{0.809}{0.007} &
\meanstd{0.041}{0.020} &
\meanstd{0.411}{0.008} &
\meanstd{\textbf{0.386}}{0.105} \\

FAVoR &
\meanstd{\textbf{0.441}}{0.021} &
\meanstd{\textbf{0.417}}{0.028} &
\meanstd{\textbf{0.889}}{0.006} &
\meanstd{\textbf{0.186}}{0.008} &
\meanstd{\textbf{0.821}}{0.007} &
\meanstd{\textbf{0.062}}{0.017} &
\meanstd{\textbf{0.413}}{0.008} &
\meanstd{0.308}{0.066} \\

Local-only (ref.) &
\meanstd{0.555}{0.006} &
\meanstd{0.509}{0.006} &
\meanstd{0.926}{0.011} &
\meanstd{0.147}{0.018} &
\meanstd{0.886}{0.003} &
\meanstd{0.154}{0.006} &
\meanstd{0.412}{0.008} &
\meanstd{0.406}{0.042} \\
\bottomrule
\end{tabularx}

\caption{K=100 BlogText scale stress test over three seeds.
Values are mean $\pm$ sample standard deviation.
Bold marks the best collaborative method; local-only is a
non-collaborative style-retention reference. Verification AUC/EER
uses same-author versus different-author generated-continuation
pairs in the K=100 ASCE space.}
\label{tab:k100-scale-stress}
\end{table*}

\subsection{Text-Volume and Author-Imbalance Sensitivity}
\label{subsec:data_scale_imbalance}

We evaluate FAVoR and Ditto over two seeds using balanced nested subsets and fixed per-client compute at 25\%, 50\%, and 100\% of a $904{,}378$-unit BlogText reference budget.
Each allocated word unit is defined by a fixed English-word regular expression; tokenizer-token counts are used separately for author filtering.
Table~\ref{tab:text-volume-sweep} shows that FAVoR retains higher Gen--Gen AUC and lower EER across the four-fold budget range, while BERTScore differences remain small.
Because these runs use rematerialized balanced subsets and a compute-normalized schedule, they support matched within-sweep comparisons rather than numerical reproduction of Table~\ref{tab:main_federated_results}.

\begin{table*}[t]
\centering
\footnotesize
\setlength{\tabcolsep}{5pt}
\renewcommand{\arraystretch}{1.06}
\begin{tabular}{lccc}
\toprule
Allocated word-unit budget & \makecell{Gen--Gen AUC\\FAVoR / Ditto} & \makecell{EER\\FAVoR / Ditto} & \makecell{BERTScore\\FAVoR / Ditto} \\
\midrule
25\% ($226{,}095$)  & 0.755 / 0.685 & 0.300 / 0.373 & 0.8141 / 0.8140 \\
50\% ($452{,}189$)  & 0.757 / 0.706 & 0.305 / 0.336 & 0.8140 / 0.8147 \\
100\% ($904{,}378$) & 0.753 / 0.699 & 0.316 / 0.360 & 0.8149 / 0.8153 \\
\bottomrule
\end{tabular}
\caption{Compute-normalized text-volume sweep. Values are means over two training seeds.}
\label{tab:text-volume-sweep}
\end{table*}

We also construct Balanced, Moderate, and Strong allocations over the same 50 authors, each containing exactly $904{,}378$ allocated word units, while keeping evaluation data and the per-client compute budget fixed.
Table~\ref{tab:imbalance-sweep} reports the corresponding two-seed means.
FAVoR retains higher AUC and lower EER in all six matched seed-by-condition comparisons; its mean AUC changes from $0.753$ to $0.738$, and absolute mean BERTScore differences remain at most $0.0018$.
This supports robustness of the relative retention advantage within the tested range, not invariance to imbalance.

\begin{table*}[t]
\centering
\footnotesize
\setlength{\tabcolsep}{4.5pt}
\renewcommand{\arraystretch}{1.06}
\begin{tabular}{lccc}
\toprule
Allocation & \makecell{Gen--Gen AUC\\FAVoR / Ditto ($\Delta$)} & \makecell{EER\\FAVoR / Ditto ($\Delta$)} & \makecell{BERTScore\\FAVoR / Ditto} \\
\midrule
Balanced & 0.753 / 0.699 ($+0.054$) & 0.316 / 0.360 ($-0.044$) & 0.8149 / 0.8153 \\
Moderate & 0.740 / 0.695 ($+0.045$) & 0.340 / 0.364 ($-0.024$) & 0.8143 / 0.8147 \\
Strong   & 0.738 / 0.702 ($+0.036$) & 0.327 / 0.360 ($-0.033$) & 0.8154 / 0.8136 \\
\bottomrule
\end{tabular}
\caption{Controlled author-data imbalance analysis. Values are means over two training seeds and $\Delta=\text{FAVoR}-\text{Ditto}$.}
\label{tab:imbalance-sweep}
\end{table*}

\subsection{Cold-Start New-Author Details}
\label{subsec:experiment_cold_start_details}

Table~\ref{tab:cold-start-full} reports the full cold-start new-author evaluation.
The shared adapter is trained on 40 federated training authors and evaluated after private $K$-shot adaptation on 10 held-out authors.
Author-separability metrics use the leakage-free A$_{\mathrm{train}}$-only ASCE encoder (Section~\ref{subsec:experiment_cold_start}) and held-out-author prototypes built only from the $K$ adaptation examples.

\begin{table*}[t]
\centering
\scriptsize
\setlength{\tabcolsep}{2.7pt}
\renewcommand{\arraystretch}{1.04}
\resizebox{\textwidth}{!}{%
\begin{tabular}{lcccccccc}
\toprule
Method & $K$ & \makecell{Sep.\\Acc. $\uparrow$} & \makecell{Macro-\\F1 $\uparrow$} &
\makecell{Target\\Sim. $\uparrow$} & \makecell{Copy-5\\$\downarrow$} &
\makecell{SemSim\\$\uparrow$} & \makecell{BERTScore\\$\uparrow$} &
\makecell{MAUVE\\$\uparrow$} \\
\midrule
Base model, no adaptation & 8 & 0.175 & 0.210 & 0.115 & 0.000 & 0.282 & 0.806 & 0.042 \\
FAVoR shared init + private residual (ASCE) & 8 & 0.538 & 0.555 & 0.353 & 0.154 & 0.304 & 0.793 & 0.071 \\
FAVoR shared init + private residual (no ASCE) & 8 & \textbf{0.641} & \textbf{0.629} & \textbf{0.363} & 0.128 & 0.301 & 0.792 & 0.078 \\
Local-only + ASCE from base & 8 & 0.400 & 0.377 & 0.261 & 0.050 & \textbf{0.331} & 0.807 & \textbf{0.115} \\
Local-only from base & 8 & 0.425 & 0.428 & 0.259 & 0.075 & 0.319 & 0.806 & 0.078 \\
Random shared init + private residual & 8 & 0.308 & 0.278 & 0.201 & 0.026 & 0.307 & \textbf{0.809} & 0.069 \\
Shared only, no private residual & 8 & 0.150 & 0.166 & 0.191 & \textbf{0.000} & 0.279 & 0.792 & 0.085 \\
\midrule
Base model, no adaptation & 16 & 0.175 & 0.150 & 0.106 & \textbf{0.000} & 0.282 & 0.806 & 0.042 \\
FAVoR shared init + private residual (ASCE) & 16 & 0.425 & 0.403 & 0.342 & 0.100 & 0.305 & 0.783 & 0.042 \\
FAVoR shared init + private residual (no ASCE) & 16 & 0.525 & 0.479 & \textbf{0.386} & 0.025 & 0.305 & 0.785 & \textbf{0.092} \\
Local-only + ASCE from base & 16 & 0.474 & 0.457 & 0.318 & \textbf{0.000} & 0.308 & 0.804 & 0.065 \\
Local-only from base & 16 & \textbf{0.553} & \textbf{0.562} & 0.326 & 0.105 & \textbf{0.313} & 0.806 & 0.048 \\
Random shared init + private residual & 16 & 0.325 & 0.320 & 0.214 & 0.025 & 0.299 & \textbf{0.809} & 0.065 \\
Shared only, no private residual & 16 & 0.150 & 0.170 & 0.189 & \textbf{0.000} & 0.279 & 0.792 & 0.085 \\
\bottomrule
\end{tabular}%
}
\caption{Full cold-start results for unseen authors. The shared adapter is trained on 40 federated training authors and transferred to 10 held-out authors; author-separability metrics use the leakage-free A$_{\mathrm{train}}$-only ASCE encoder and held-out-author prototypes built from the same $K$ adaptation examples.}
\label{tab:cold-start-full}
\end{table*}

To characterize the transition beyond the original $K=8$ and $K=16$ endpoints, we freeze one federated shared checkpoint and rerun the no-ASCE FAVoR and local-only variants over nested $K\in\{4,8,12,16,24\}$ support sets with three seeds and the same 10 unseen authors.
In addition to the original $K$-shot prototypes, this diagnostic uses fixed 32-text author prototypes disjoint from adaptation and test data, separating adaptation effects from prototype-quality changes.
Let $\Delta=M_{\mathrm{FAVoR}}-M_{\mathrm{Local}}$, and let $m$ denote target-author similarity minus shard-negative similarity.

\begin{table}[H]
\centering
\footnotesize
\setlength{\tabcolsep}{3.5pt}
\renewcommand{\arraystretch}{1.06}
\begin{tabular}{ccc}
\toprule
$K$ & Fixed $\Delta$ Acc. [95\% CI] & Fixed $\Delta m$ [95\% CI] \\
\midrule
4  & $+0.086$ $[-0.081,\ 0.244]$ & $+0.046$ $[-0.017,\ 0.103]$ \\
8  & $+0.119$ $[0.050,\ 0.203]$  & $+0.050$ $[0.003,\ 0.096]$ \\
12 & $+0.064$ $[-0.017,\ 0.150]$ & $+0.036$ $[0.006,\ 0.065]$ \\
16 & $+0.053$ $[-0.022,\ 0.133]$ & $+0.032$ $[-0.008,\ 0.073]$ \\
24 & $-0.044$ $[-0.172,\ 0.067]$ & $+0.027$ $[-0.027,\ 0.076]$ \\
\bottomrule
\end{tabular}
\caption{Nested cold-start $K$-sweep with fixed 32-text evaluation prototypes. Intervals are nominal author-cluster 95\% bootstrap confidence intervals.}
\label{tab:cold-start-nested-k}
\end{table}

Table~\ref{tab:cold-start-nested-k} shows that both the $K$-shot and fixed-prototype diagnostics give the clearest FAVoR advantage at $K=8$, diminishing differences by $K=16$, and no reliable winner at $K=16$--$24$.
The result supports a diminishing shared-initialization advantage rather than a universal crossover threshold.

\subsection{Human 2AFC Audit}
\label{subsec:human_2afc_audit}

We conduct a small human two-alternative forced-choice (2AFC) audit as a perceptual sanity check for author-style retention.
For each item, annotators are shown two anonymized source-based author profiles and one candidate continuation, and are asked to choose which profile better matches the candidate.
Annotators also provide a confidence score from 1 to 5.
The task is designed to test whether an author-specific signal in a continuation is human-perceivable under a controlled attribution setting; it is not a preference study and does not ask annotators to judge overall writing quality, fluency, or authenticity.

The audit set contains 50 items randomly sampled within source categories: 20 Ditto continuations, 20 FAVoR continuations, and 10 held-out author-written continuations.
Each item is annotated by three annotators, giving 150 total votes.
We aggregate results both at the vote level, treating each annotator's decision as one observation, and at the item level, using the majority decision across the three annotators.

\paragraph{Participant instructions.}
The annotation interface was a text-only form with two author-profile boxes, one candidate continuation, a forced-choice response, a confidence response, and an optional comment field.
No screenshots contained additional instructions beyond the text reproduced here.
Annotators were shown the following instructions before the task:
\begin{quote}
\small
This pack is for a 2-author forced-choice style attribution task.
Annotator task:
1. Read Author A's source excerpts.
2. Read Author B's source excerpts.
3. Read the candidate text.
4. Choose which author the candidate better matches in writing style.
5. Enter confidence from 1 to 5.
\end{quote}

\begin{table}[t]
\centering
\footnotesize
\setlength{\tabcolsep}{0pt}
\renewcommand{\arraystretch}{1.08}
\begin{tabular*}{\columnwidth}{@{\extracolsep{\fill}}lccccc@{}}
\toprule
Source & Items & Votes & \makecell{Vote\\acc.} & \makecell{Maj.\\acc.} & \makecell{Mean\\conf.} \\
\midrule
Ditto & 20 & 60 & 68.3\% & 70.0\% & 2.73 \\
FAVoR & 20 & 60 & 91.7\% & 95.0\% & 3.79 \\
Human ref. & 10 & 30 & 93.3\% & 90.0\% & 3.50 \\
\midrule
Overall & 50 & 150 & 82.7\% & 84.0\% & 3.31 \\
\bottomrule
\end{tabular*}
\caption{Small human 2AFC audit. Vote accuracy is over individual annotator votes; majority accuracy is after three-vote item aggregation.}
\label{tab:human_2afc_audit}
\end{table}

Table~\ref{tab:human_2afc_audit} summarizes the audit results.
Overall vote-level accuracy is 82.7\%, and item-level majority accuracy is 84.0\%.
On the audited generated continuations, FAVoR continuations are attributed to the intended author more often than Ditto continuations, with 91.7\% versus 68.3\% vote-level accuracy and 95.0\% versus 70.0\% majority accuracy.
Held-out human continuations provide a reference row and reach 93.3\% vote-level accuracy.
Inter-annotator agreement is moderate for this subjective authorship task: mean raw pairwise agreement is 0.787 and Fleiss' $\kappa$ is 0.558.
At the item level, all 50 items have a non-tie majority; 34 items have unanimous 3--0 majorities and 16 items have 2--1 majorities.
This audit supports the interpretation that FAVoR preserves an author-attributable signal that is perceptible to human annotators in a 2AFC attribution setting.
However, it has two limitations.
First, the audit is small, with 50 items and three annotators.
Second, because annotators see author profiles, their decisions may reflect topical, biographical, or content cues in addition to style.
We therefore use the audit only as complementary perceptual evidence alongside the main automatic author-retention metrics and ASCE-independent external verifiers, rather than as a standalone human preference or authenticity evaluation.

\subsection{Reference-Free Quality and Voice-Match Judges}
\label{subsec:experiment_llm_judges}

We supplement reference-based utility metrics with a method-blinded evaluation using the fixed \texttt{qwen3.7-plus-2026-05-26} snapshot.
For generic quality, the judge receives only a prompt prefix and one continuation---not the gold continuation, author history, or method identity---and scores coherence, fluency, specificity, non-repetition, and overall quality from 1 to 5.
The matched design uses 50 authors and three generation seeds, with one common prompt for all six collaborative methods in each author--seed group, yielding 150 matched groups and 900 outputs.
Each output is judged twice, for 1,800 judgments with no missing or failed calls; the two scores are averaged before paired analysis, and exact repeated-score agreement on overall quality is $92.1\%$.

Confidence intervals use 10,000 author-cluster bootstrap resamples that resample the 50 authors while retaining all three seeds for each selected author.
Two-sided raw $p$ values come from the same bootstrap distribution and are Holm-adjusted across the five FAVoR--baseline comparisons.
Table~\ref{tab:qwen-generic-quality} shows that all five confidence intervals include zero and no generic-quality difference is significant after correction.

\begin{table*}[t]
\centering
\footnotesize
\setlength{\tabcolsep}{5pt}
\renewcommand{\arraystretch}{1.06}
\begin{tabular}{lcccc}
\toprule
Method & Mean quality & FAVoR $-$ method & Unadjusted 95\% CI & Holm-adjusted $p$ \\
\midrule
FedAvg  & 2.270 & $-0.073$ & $[-0.310,\ 0.170]$ & 1.000 \\
FedProx & 2.213 & $-0.017$ & $[-0.267,\ 0.243]$ & 1.000 \\
pFedMe  & 2.133 & $+0.063$ & $[-0.217,\ 0.323]$ & 1.000 \\
Ditto   & 2.370 & $-0.173$ & $[-0.443,\ 0.103]$ & 1.000 \\
FedDPA  & 2.337 & $-0.140$ & $[-0.380,\ 0.097]$ & 1.000 \\
FAVoR   & 2.197 & -- & -- & -- \\
\bottomrule
\end{tabular}
\caption{Reference-free overall-quality judging on the 1--5 scale. Differences are computed from unrounded output-level means.}
\label{tab:qwen-generic-quality}
\end{table*}

In a separate method-blinded pairwise FAVoR--Ditto author-voice comparison, the judge prefers FAVoR in $61.2\%$ of non-ties (95\% CI $[0.518,0.702]$, author-clustered $p=0.022$).
Thus, the reference-free evidence does not establish generic-quality superiority or equivalence, but the voice-match result supports FAVoR when author-style retention is the primary objective.

\subsection{External Verifier Robustness}
\label{subsec:experiment_external_verifiers}

The ASCE-independent verifiers are StyleDistance embeddings, all-MiniLM-L6-v2 sentence embeddings, and hand-engineered stylometric features.
The stylometric representation includes character and word $n$-grams, function-word frequencies, and punctuation and structure statistics, following \citet{stamatatos2009survey}.
Generated-to-source cosine verification uses generated/source pairs from the same author as positives and source texts from different authors in the same usable metadata-topic bucket as negatives.
AUC is the area under the receiver operating characteristic curve, and EER is the equal error rate for these same-author versus different-author pairs.

Table~\ref{tab:phase1-main-external-verification} reports compact external-verification averages; Table~\ref{tab:phase1-external-evaluator-details} gives the full evaluator breakdown.
Table~\ref{tab:phase1-external-agreement} summarizes cross-evaluator agreement for the central FAVoR-versus-Ditto comparison.
Table~\ref{tab:phase1-topic-audit} documents topic recovery, the effective generated-pair negative mode, and same-topic hard-negative feasibility for the verification protocol.
Table~\ref{tab:phase1-legacy-comparison} provides an additional compact BlogText metric view.
Source AUC is a verifier-validity check on held-out human source-text pairs, while generated-text columns evaluate method outputs under the same-topic cross-author negatives.
The \texttt{indUnk} metadata bucket is retained in the evaluation roster but treated as unavailable for same-topic negative sampling.
Bold and underline mark the best and second-best values, with lower better for EER.

\begin{table*}[t]
\centering
\scriptsize
\setlength{\tabcolsep}{2.5pt}
\resizebox{\textwidth}{!}{%
\begin{tabular}{llcccc}
\toprule
Method & Evaluator & Source AUC & Gen$\rightarrow$Src AUC & Gen$\rightarrow$Src EER & Gen$\rightarrow$Gen AUC \\
\midrule
FedProx & StyleDistance & 0.705 $\pm$ 0.000 & 0.724 $\pm$ 0.011 & 0.334 $\pm$ 0.006 & 0.654 $\pm$ 0.021 \\
FedProx & MiniLM & 0.635 $\pm$ 0.000 & 0.631 $\pm$ 0.002 & 0.414 $\pm$ 0.003 & 0.532 $\pm$ 0.005 \\
FedProx & Stylometry & 0.619 $\pm$ 0.000 & 0.651 $\pm$ 0.003 & 0.385 $\pm$ 0.005 & 0.573 $\pm$ 0.021 \\
\midrule
Pooled PEFT & StyleDistance & 0.705 $\pm$ 0.000 & 0.736 $\pm$ 0.006 & 0.327 $\pm$ 0.004 & 0.691 $\pm$ 0.016 \\
Pooled PEFT & MiniLM & 0.635 $\pm$ 0.000 & 0.650 $\pm$ 0.001 & 0.406 $\pm$ 0.005 & 0.574 $\pm$ 0.004 \\
Pooled PEFT & Stylometry & 0.619 $\pm$ 0.000 & 0.660 $\pm$ 0.008 & 0.384 $\pm$ 0.008 & 0.562 $\pm$ 0.022 \\
\midrule
Ditto & StyleDistance & 0.705 $\pm$ 0.000 & 0.748 $\pm$ 0.004 & 0.318 $\pm$ 0.006 & 0.720 $\pm$ 0.009 \\
Ditto & MiniLM & 0.635 $\pm$ 0.000 & 0.650 $\pm$ 0.005 & 0.401 $\pm$ 0.005 & 0.594 $\pm$ 0.013 \\
Ditto & Stylometry & 0.619 $\pm$ 0.000 & 0.683 $\pm$ 0.011 & 0.363 $\pm$ 0.008 & 0.656 $\pm$ 0.034 \\
\midrule
FedDPA\textsuperscript{\dag} & StyleDistance & 0.705 $\pm$ 0.000 & 0.747 $\pm$ 0.005 & 0.318 $\pm$ 0.003 & 0.725 $\pm$ 0.011 \\
FedDPA\textsuperscript{\dag} & MiniLM & 0.635 $\pm$ 0.000 & 0.659 $\pm$ 0.012 & 0.392 $\pm$ 0.012 & 0.573 $\pm$ 0.027 \\
FedDPA\textsuperscript{\dag} & Stylometry & 0.619 $\pm$ 0.000 & 0.692 $\pm$ 0.008 & 0.351 $\pm$ 0.008 & 0.652 $\pm$ 0.007 \\
\midrule
Local-only + ASCE & StyleDistance & 0.705 $\pm$ 0.000 & \underline{0.768 $\pm$ 0.006} & \underline{0.301 $\pm$ 0.002} & \underline{0.768 $\pm$ 0.004} \\
Local-only + ASCE & MiniLM & 0.635 $\pm$ 0.000 & \textbf{0.685 $\pm$ 0.006} & \textbf{0.374 $\pm$ 0.008} & \textbf{0.645 $\pm$ 0.023} \\
Local-only + ASCE & Stylometry & 0.619 $\pm$ 0.000 & \underline{0.708 $\pm$ 0.003} & \underline{0.344 $\pm$ 0.007} & \textbf{0.685 $\pm$ 0.010} \\
\midrule
FAVoR & StyleDistance & 0.705 $\pm$ 0.000 & \textbf{0.771 $\pm$ 0.006} & \textbf{0.295 $\pm$ 0.003} & \textbf{0.773 $\pm$ 0.007} \\
FAVoR & MiniLM & 0.635 $\pm$ 0.000 & \underline{0.675 $\pm$ 0.010} & \underline{0.378 $\pm$ 0.006} & \underline{0.624 $\pm$ 0.021} \\
FAVoR & Stylometry & 0.619 $\pm$ 0.000 & \textbf{0.714 $\pm$ 0.011} & \textbf{0.342 $\pm$ 0.011} & \underline{0.678 $\pm$ 0.020} \\
\bottomrule
\end{tabular}%
}
\caption{External-verifier breakdown.}
\label{tab:phase1-external-evaluator-details}
\end{table*}

\begin{table*}[t]
\centering
\footnotesize
\setlength{\tabcolsep}{1.7pt}
\renewcommand{\arraystretch}{1.14}

\begin{tabularx}{\textwidth}{
@{}>{\raggedright\arraybackslash}Xcccccccc@{}}
\toprule
Method &
\makecell{Author\\Acc. $\uparrow$} &
\makecell{Macro-\\F1 $\uparrow$} &
\makecell{Centroid\\$\uparrow$} &
\makecell{Silhouette\\$\uparrow$} &
\makecell{Asst.\\$\downarrow$} &
\makecell{BERTScore\\$\uparrow$} &
\makecell{SemSim\\$\uparrow$} &
\makecell{MAUVE\\$\uparrow$} \\
\midrule
FedProx &
\meanstd{0.174}{0.008} &
\meanstd{0.161}{0.009} &
\meanstd{0.431}{0.009} &
\meanstd{-0.199}{0.004} &
\meanstd{0.264}{0.041} &
\meanstd{\textbf{0.817}}{0.002} &
\meanstd{0.353}{0.005} &
\meanstd{0.200}{0.005} \\

Pooled PEFT &
\meanstd{0.254}{0.021} &
\meanstd{0.232}{0.027} &
\meanstd{0.537}{0.005} &
\meanstd{-0.184}{0.030} &
\meanstd{\textbf{0.219}}{0.017} &
\meanstd{0.802}{0.001} &
\meanstd{0.359}{0.002} &
\meanstd{0.389}{0.059} \\

Ditto &
\meanstd{0.278}{0.011} &
\meanstd{0.255}{0.002} &
\meanstd{0.546}{0.004} &
\meanstd{-0.097}{0.026} &
\meanstd{0.328}{0.016} &
\meanstd{\underline{0.815}}{0.001} &
\meanstd{0.366}{0.005} &
\meanstd{0.378}{0.074} \\

Local-only + ASCE &
\meanstd{\underline{0.402}}{0.026} &
\meanstd{\textbf{0.366}}{0.016} &
\meanstd{\textbf{0.643}}{0.012} &
\meanstd{\textbf{-0.037}}{0.025} &
\meanstd{\underline{0.239}}{0.015} &
\meanstd{0.813}{0.001} &
\meanstd{\underline{0.375}}{0.009} &
\meanstd{\textbf{0.486}}{0.053} \\

FAVoR &
\meanstd{\textbf{0.403}}{0.004} &
\meanstd{\underline{0.365}}{0.009} &
\meanstd{\underline{0.624}}{0.006} &
\meanstd{\underline{-0.039}}{0.022} &
\meanstd{0.260}{0.004} &
\meanstd{0.815}{0.000} &
\meanstd{\textbf{0.380}}{0.001} &
\meanstd{\underline{0.431}}{0.026} \\
\bottomrule
\end{tabularx}

\caption{Additional compact BlogText metric summary.}
\label{tab:phase1-legacy-comparison}
\end{table*}

\begin{table}[t]
\centering
\footnotesize
\setlength{\tabcolsep}{3pt}
\renewcommand{\arraystretch}{1.07}

\begin{tabularx}{\columnwidth}{
@{}p{0.40\columnwidth}>{\raggedright\arraybackslash}X@{}}
\toprule
Audit item & Value \\
\midrule
Evaluation authors & 50 \\
Topic source & Original BlogText metadata joined by author ID \\
Topics with $\geq 3$ authors & 3 usable; 4 including all buckets \\
Largest topic bucket & Student: 13/50 (26.0\%) \\
indUnk bucket & 11/50 (22.0\%) \\
Dominant-topic fraction & mean 1.0; min 1.0 \\
Effective hard-negative mode & Same-topic cross-author \\
Protocol A same-prompt feasible & False \\
Protocol A same-topic negatives & 2818--2850 \\
Protocol C same-prompt feasible & False \\
Protocol C same-topic negatives & 1409--1425 \\
External-evaluator agreement & 3/3 support FAVoR $>$ Ditto \\
\bottomrule
\end{tabularx}

\caption{Topic and negative-sampling summary.}
\label{tab:phase1-topic-audit}
\end{table}

\begin{table*}[t]
\centering
\scriptsize
\setlength{\tabcolsep}{4pt}
\resizebox{\textwidth}{!}{%
\begin{tabular}{llccc}
\toprule
Comparison & Metric & Support & Fraction & Supporting evaluators \\
\midrule
FAVoR $>$ Ditto & Gen$\rightarrow$Src primary AUC & 3/3 & 1.0 & minilm, styledistance, stylometry \\
\bottomrule
\end{tabular}%
}
\caption{Cross-evaluator agreement.}
\label{tab:phase1-external-agreement}
\end{table*}

\subsection{Author-Disjoint ASCE Training and Evaluation Details}
\label{subsec:experiment_asce_leakage_free_audit}

The production BlogText ASCE encoder used for local style alignment and ASCE-space evaluation is trained centrally before FL on 50 BlogText authors disjoint from the 50-author FL roster.
It uses 4,959 training texts and 742 validation texts, achieving 0.802 validation accuracy and 0.673 macro-F1.
It does not observe FL training texts, held-out prompts, generated continuations, or any main-roster author identity, and the manifest records zero author overlap with the FL roster.
After training, its classifier head and learned author prototypes are discarded and the encoder is frozen.
ASCE-aligned methods use this frozen encoder during training to construct local style-alignment targets, and all BlogText ASCE-space author-style metrics use the same author-disjoint encoder.

The encoder backbone is \texttt{distilroberta-base}: a six-layer DistilRoBERTa with hidden size 768, 12 attention heads, and feed-forward intermediate size 3,072.
We apply attention-mask mean pooling to its final-layer token embeddings, followed by dropout with rate 0.1, a linear projection from 768 to 256 dimensions, \texttt{LayerNorm(256)}, and $\ell_2$ normalization to obtain the 256-dimensional ASCE embedding.
During training, this embedding is connected to a normalized ArcFace author-classification head.

For completeness, ASCE maps a text $x$ to the unit-normalized embedding
\[
z(x)=\operatorname{norm}\!\left(E_{\phi}(x)\right).
\]
For a text from author $a$, let $w_a$ be the normalized
target-author classifier prototype and let $w_k$ denote a
non-target prototype. The angular-margin logits are
\[
\begin{aligned}
s_a(x)
&=
\gamma\cos\!\left(
\arccos\!\left(w_a^\top z(x)\right)+m_s
\right),\\
s_k(x)
&=
\gamma w_k^\top z(x),
\qquad k\neq a.
\end{aligned}
\]
The corresponding softmax normalizer and cross-entropy loss are
\[
\begin{aligned}
Z_a(x)
&=
\exp\!\left(s_a(x)\right)
+\sum_{k\neq a}\exp\!\left(s_k(x)\right),\\
\mathcal{L}^{\mathrm{ASCE}}(x,a)
&=
-s_a(x)+\log Z_a(x).
\end{aligned}
\]
Here $\gamma$ is the scale applied to normalized cosine logits,
$m_s$ is the angular margin, and $Z_a(x)$ is the softmax
normalizer. 
For the authorship ASCE used in our experiments, $\gamma=30.0$ and $m_s=0.35$ radians.
ASCE-A and ASCE-B use the same encoder backbone, architecture, angular-margin objective, and source domain; ASCE-B differs in its disjoint training authors and is used only for post-hoc evaluation, as detailed in Appendix~\ref{subsec:asce_b_audit}.
The LoRA/adapter settings for the generation model are reported separately in Appendix~\ref{subsec:experiment_efficiency}.

Tables~\ref{tab:asce-leakage-free-gensource} and~\ref{tab:asce-leakage-free-pairwise} give two detailed views of the author-disjoint ASCE results behind the main BlogText summary.
Table~\ref{tab:asce-leakage-free-gensource} is source-anchored: it compares each generated continuation against source-train author prototypes and gen$\rightarrow$source verification pairs.
Table~\ref{tab:asce-leakage-free-pairwise} is the expanded version of the BlogText author-style rows in Table~\ref{tab:main_federated_results}, adding pooled and diagnostic references such as Local-only + ASCE, Global ASCE, Ditto + ASCE, and FAVoR w/o ASCE.

\begin{table*}[t]
\centering
\scriptsize
\setlength{\tabcolsep}{4pt}
\resizebox{\textwidth}{!}{%
\begin{tabular}{lccccc}
\toprule
Method & Seeds & \makecell{Proto. Acc.\\$\uparrow$} & \makecell{Proto. F1\\$\uparrow$} & \makecell{Gen$\rightarrow$Src\\AUC $\uparrow$} & \makecell{Gen$\rightarrow$Src\\EER $\downarrow$} \\
\midrule
FedAvg & 3 & 0.140 $\pm$ 0.009 & 0.096 $\pm$ 0.001 & 0.651 $\pm$ 0.023 & 0.379 $\pm$ 0.008 \\
FedProx & 3 & 0.157 $\pm$ 0.011 & 0.111 $\pm$ 0.005 & 0.645 $\pm$ 0.023 & 0.387 $\pm$ 0.020 \\
pFedMe & 3 & 0.176 $\pm$ 0.014 & 0.120 $\pm$ 0.010 & 0.691 $\pm$ 0.019 & 0.363 $\pm$ 0.018 \\
Ditto & 3 & 0.219 $\pm$ 0.016 & 0.169 $\pm$ 0.012 & 0.715 $\pm$ 0.014 & 0.351 $\pm$ 0.008 \\
Pooled PEFT & 3 & 0.193 $\pm$ 0.008 & 0.148 $\pm$ 0.012 & 0.689 $\pm$ 0.012 & 0.363 $\pm$ 0.016 \\
Local-only + ASCE & 3 & \textbf{0.342 $\pm$ 0.023} & \textbf{0.292 $\pm$ 0.032} & \textbf{0.764 $\pm$ 0.006} & \underline{0.307 $\pm$ 0.013} \\
Global ASCE & 3 & 0.149 $\pm$ 0.008 & 0.112 $\pm$ 0.007 & 0.664 $\pm$ 0.024 & 0.378 $\pm$ 0.032 \\
Ditto + ASCE & 3 & 0.237 $\pm$ 0.006 & 0.173 $\pm$ 0.016 & 0.699 $\pm$ 0.020 & 0.366 $\pm$ 0.021 \\
FAVoR w/o ASCE & 3 & 0.280 $\pm$ 0.031 & 0.223 $\pm$ 0.035 & 0.741 $\pm$ 0.023 & 0.326 $\pm$ 0.029 \\
FAVoR & 3 & \underline{0.315 $\pm$ 0.022} & \underline{0.262 $\pm$ 0.031} & \underline{0.752 $\pm$ 0.026} & \textbf{0.304 $\pm$ 0.016} \\
\bottomrule
\end{tabular}%
}
\caption{Generated-to-source ASCE results with the author-disjoint encoder. The encoder is trained on BlogText authors disjoint from the main 50-author FL roster; its classifier head is not used for this table. Values are the mean $\pm$ standard deviation over seeds.}
\label{tab:asce-leakage-free-gensource}
\end{table*}

\begin{table*}[t]
\centering
\scriptsize
\setlength{\tabcolsep}{4pt}
\resizebox{\textwidth}{!}{%
\begin{tabular}{lccccc}
\toprule
Method &
\makecell{Author Acc.\\$\uparrow$} &
\makecell{Macro-F1\\$\uparrow$} &
\makecell{Silhouette\\$\uparrow$} &
\makecell{Gen-Gen AUC\\$\uparrow$} &
\makecell{Gen-Gen EER\\$\downarrow$} \\
\midrule
FedAvg & 0.199 $\pm$ 0.008 & 0.164 $\pm$ 0.015 & -0.194 $\pm$ 0.014 & 0.680 $\pm$ 0.029 & 0.371 $\pm$ 0.018 \\
FedProx & 0.174 $\pm$ 0.008 & 0.161 $\pm$ 0.009 & -0.199 $\pm$ 0.004 & 0.674 $\pm$ 0.003 & 0.365 $\pm$ 0.011 \\
pFedMe & 0.271 $\pm$ 0.047 & 0.252 $\pm$ 0.033 & -0.131 $\pm$ 0.005 & 0.739 $\pm$ 0.011 & 0.321 $\pm$ 0.014 \\
Ditto & 0.278 $\pm$ 0.011 & 0.255 $\pm$ 0.002 & -0.097 $\pm$ 0.026 & 0.763 $\pm$ 0.025 & 0.297 $\pm$ 0.018 \\
Ditto + ASCE & 0.318 $\pm$ 0.037 & 0.286 $\pm$ 0.046 & -0.115 $\pm$ 0.004 & 0.761 $\pm$ 0.010 & 0.309 $\pm$ 0.010 \\
Pooled PEFT & 0.254 $\pm$ 0.021 & 0.232 $\pm$ 0.027 & -0.184 $\pm$ 0.030 & 0.699 $\pm$ 0.020 & 0.344 $\pm$ 0.009 \\
Local-only + ASCE & \underline{0.402 $\pm$ 0.026} & \textbf{0.366 $\pm$ 0.016} & \underline{-0.037 $\pm$ 0.025} & \textbf{0.828 $\pm$ 0.017} & 0.252 $\pm$ 0.018 \\
Global ASCE & 0.161 $\pm$ 0.014 & 0.140 $\pm$ 0.012 & -0.195 $\pm$ 0.010 & 0.681 $\pm$ 0.012 & 0.368 $\pm$ 0.009 \\
FAVoR w/o ASCE & 0.377 $\pm$ 0.036 & 0.331 $\pm$ 0.033 & \textbf{-0.036 $\pm$ 0.008} & 0.816 $\pm$ 0.006 & \underline{0.247 $\pm$ 0.010} \\
FAVoR & \textbf{0.403 $\pm$ 0.004} & \underline{0.365 $\pm$ 0.009} & -0.039 $\pm$ 0.022 & \underline{0.824 $\pm$ 0.006} & \textbf{0.245 $\pm$ 0.007} \\
\bottomrule
\end{tabular}%
}
\caption{Expanded BlogText author-style results under the author-disjoint ASCE encoder. This table expands the BlogText author-style rows in Table~\ref{tab:main_federated_results} with pooled and diagnostic references. Accuracy and F1 use source-train prototypes, while Gen-Gen AUC/EER measure same-author versus different-author verification among generated continuations. Values are mean $\pm$ standard deviation over seeds.}
\label{tab:asce-leakage-free-pairwise}
\end{table*}

Together, these two views document author-style retention in the same author-disjoint ASCE space used by the BlogText training and evaluation protocol.
In the gen$\rightarrow$source view, local-only remains the strongest non-collaborative reference, while FAVoR is the strongest collaborative method: it ranks second overall on prototype recovery and AUC, and best overall on EER.
In the expanded generated-text view, FAVoR again leads the collaborative methods on author recovery and verification EER, and remains close to the local-only reference on F1 and AUC.
The FAVoR variant without ASCE is also competitive, indicating that residual personalization carries substantial author signal; adding ASCE improves recovery and verification while leaving silhouette essentially tied with the local-only reference.

\subsection{Independent ASCE-B Re-evaluation}
\label{subsec:asce_b_audit}

To reduce same-checkpoint circularity, we train an independent ASCE-B from the same backbone on 50 additional BlogText authors with zero overlap with either the ASCE-A training authors or the FL roster.
ASCE-B uses 6,460 training texts and 982 validation texts and reaches validation accuracy/macro-F1 of $0.844/0.784$.
It is used only for post-hoc evaluation: archived generations and evaluation pairs from the same three seeds are fixed while embeddings, source prototypes, and verification scores are recomputed.

\begin{table}[t]
\centering
\footnotesize
\setlength{\tabcolsep}{4pt}
\renewcommand{\arraystretch}{1.06}
\begin{tabular}{lcc}
\toprule
Collaborative method & \makecell{Source-prototype\\Acc.} & \makecell{Gen$\rightarrow$Src\\AUC} \\
\midrule
FAVoR          & \textbf{0.403 $\pm$ 0.017} & \textbf{0.771 $\pm$ 0.011} \\
FAVoR w/o ASCE & 0.347 $\pm$ 0.016 & \underline{0.766 $\pm$ 0.009} \\
Ditto          & 0.214 $\pm$ 0.022 & 0.723 $\pm$ 0.014 \\
FedDPA         & \underline{0.242 $\pm$ 0.034} & 0.713 $\pm$ 0.018 \\
pFedMe         & 0.195 $\pm$ 0.003 & 0.707 $\pm$ 0.030 \\
FedAvg         & 0.137 $\pm$ 0.028 & 0.663 $\pm$ 0.012 \\
FedProx        & 0.143 $\pm$ 0.006 & 0.657 $\pm$ 0.020 \\
\bottomrule
\end{tabular}
\caption{Post-hoc evaluation under independently trained ASCE-B. Archived generations and pair construction are held fixed; values are mean $\pm$ sample standard deviation over three seeds.}
\label{tab:asce-b-audit}
\end{table}

Table~\ref{tab:asce-b-audit} shows that FAVoR remains highest under ASCE-B, and FAVoR without ASCE alignment exceeds all non-FAVoR collaborative baselines.
Thus, same-checkpoint circularity cannot solely explain the gain, although ASCE-B still shares ASCE-A's architecture, objective, and source domain.

\subsection{ASCE Evaluator-Scale and Cross-Domain Transfer}
\label{subsec:asce_scale_transfer}

We train independent ASCE checkpoints using $K_{\mathrm{ASCE}}\in\{10,25,50\}$ author-disjoint BlogText authors.
The sensitivity ASCE-50 is not the production checkpoint despite using the same 50-author, 4,959/742 train/validation split.
Archived FL outputs, prototype source texts, evaluation splits, and pair construction are fixed, and embeddings and prototypes are recomputed in each checkpoint space.
We then apply each frozen checkpoint to 50 Mythos authors unseen during ASCE training.

\begin{table*}[t]
\centering
\footnotesize
\setlength{\tabcolsep}{3.6pt}
\renewcommand{\arraystretch}{1.06}
\begin{tabular}{cccccc}
\toprule
\makecell{ASCE training\\$K_{\mathrm{ASCE}}$} & Examples & \makecell{BlogText Proto-Acc.\\FAVoR / Ditto / FedAvg} & \makecell{BlogText Gen$\rightarrow$Src AUC\\FAVoR / Ditto / FedAvg} & \makecell{Mythos Proto-Acc.\\{[95\% CI]}} & \makecell{Mythos AUC\\{[95\% CI]}} \\
\midrule
10 & 1,020 & 0.322 / 0.211 / 0.125 & 0.726 / 0.696 / 0.648 & 0.203 $[0.147,0.263]$ & 0.647 $[0.618,0.677]$ \\
25 & 2,749 & 0.378 / 0.241 / 0.147 & 0.757 / 0.709 / 0.659 & 0.277 $[0.217,0.337]$ & 0.712 $[0.680,0.744]$ \\
50 & 4,959 & 0.293 / 0.199 / 0.147 & 0.754 / 0.685 / 0.631 & 0.220 $[0.160,0.283]$ & 0.650 $[0.623,0.680]$ \\
\bottomrule
\end{tabular}
\caption{ASCE evaluator-scale sensitivity on fixed BlogText outputs and evaluation-only transfer to unseen Mythos authors.}
\label{tab:asce-scale-transfer}
\end{table*}

Table~\ref{tab:asce-scale-transfer} shows that, within BlogText, the ordering FAVoR $>$ Ditto $>$ FedAvg is preserved for both metrics at every $K_{\mathrm{ASCE}}$.
Across Mythos, all prototype accuracies exceed 50-way chance accuracy ($0.020$), and all AUC intervals exclude $0.5$.
These results show nontrivial representation transfer within the tested settings, not a minimum ASCE author count, prototype-free zero-shot attribution, or universal out-of-domain robustness.

\subsection{External Domain Validation}
\label{subsec:experiment_mythos_external}

Table~\ref{tab:mythos-reddit-external-main} reports the Mythos-Reddit external comparison for the collaborative methods.
Table~\ref{tab:mythos-reddit-external-details} reports the diagnostic version of the external evaluation, including signed-gap metrics and assistant-phrase diagnostics.
The completed validation uses 50 r/WritingPrompts authors with 300 balanced held-out continuations per run.
On Mythos-Reddit, FAVoR is again the strongest collaborative federated method on the style-separability columns, suggesting that the retention trend is not specific to BlogText.

\begin{table*}[t]
\centering
\footnotesize
\setlength{\tabcolsep}{1.6pt}
\renewcommand{\arraystretch}{1.14}

\begin{tabularx}{\textwidth}{
@{}>{\raggedright\arraybackslash}Xcccccccc@{}}
\toprule
Method &
\makecell{Sep.\\Acc. $\uparrow$} &
\makecell{Macro-\\F1 $\uparrow$} &
\makecell{Centroid\\$\uparrow$} &
\makecell{Silhouette\\$\uparrow$} &
\makecell{AUC\\$\uparrow$} &
\makecell{EER\\$\downarrow$} &
\makecell{BERTScore\\$\uparrow$} &
\makecell{SemSim\\$\uparrow$} \\
\midrule
FedAvg &
\meanstd{0.134}{0.023} & \meanstd{0.115}{0.023} &
\meanstd{0.455}{0.083} & \meanstd{-0.193}{0.024} &
\meanstd{0.658}{0.024} & \meanstd{0.386}{0.026} &
\meanstd{0.823}{0.000} & \meanstd{0.482}{0.009} \\

FedProx &
\meanstd{0.141}{0.024} & \meanstd{0.113}{0.017} &
\meanstd{0.463}{0.096} & \meanstd{-0.172}{0.016} &
\meanstd{0.663}{0.029} & \meanstd{0.377}{0.026} &
\meanstd{0.823}{0.000} & \meanstd{0.478}{0.008} \\

pFedMe &
\meanstd{0.324}{0.010} & \meanstd{0.290}{0.015} &
\meanstd{0.654}{0.040} & \meanstd{-0.054}{0.008} &
\meanstd{0.762}{0.005} & \meanstd{0.305}{0.005} &
\meanstd{0.823}{0.000} & \meanstd{0.477}{0.002} \\

Ditto &
\meanstd{0.392}{0.022} & \meanstd{0.359}{0.029} &
\meanstd{0.685}{0.030} & \meanstd{-0.018}{0.007} &
\meanstd{0.818}{0.002} & \meanstd{0.253}{0.002} &
\meanstd{0.825}{0.000} & \meanstd{0.487}{0.006} \\

FedDPA\textsuperscript{\dag} &
\meanstd{0.357}{0.035} & \meanstd{0.317}{0.033} &
\meanstd{0.667}{0.048} & \meanstd{-0.053}{0.021} &
\meanstd{0.803}{0.013} & \meanstd{0.268}{0.010} &
\meanstd{0.825}{0.000} & \meanstd{0.488}{0.005} \\

FAVoR &
\meanstd{0.561}{0.024} & \meanstd{0.521}{0.029} &
\meanstd{0.771}{0.021} & \meanstd{0.093}{0.017} &
\meanstd{0.900}{0.012} & \meanstd{0.175}{0.011} &
\meanstd{0.826}{0.000} & \meanstd{0.494}{0.005} \\
\bottomrule
\end{tabularx}

\caption{Mythos-Reddit external validation.}
\label{tab:mythos-reddit-external-main}
\end{table*}

\begin{table*}[t]
\centering
\footnotesize
\setlength{\tabcolsep}{1.2pt}
\renewcommand{\arraystretch}{1.14}

\begin{tabularx}{\textwidth}{
@{}>{\raggedright\arraybackslash}Xccccccccc@{}}
\toprule
Method &
\makecell{Sep.\\Acc. $\uparrow$} &
\makecell{Macro-\\F1 $\uparrow$} &
\makecell{Centroid\\$\uparrow$} &
\makecell{Silhouette\\$\uparrow$} &
\makecell{Phrase\\$\downarrow$} &
\makecell{BERTScore\\$\uparrow$} &
\makecell{SemSim\\$\uparrow$} &
\makecell{Gap-Silh.\\$\uparrow$} &
\makecell{Gap-Cent.\\$\uparrow$} \\
\midrule
FedAvg &
\meanstd{0.134}{0.023} & \meanstd{0.115}{0.023} &
\meanstd{0.455}{0.083} & \meanstd{-0.193}{0.024} &
\meanstd{1.434}{0.367} & \meanstd{0.823}{0.000} &
\meanstd{0.482}{0.009} & \meanstd{-0.237}{0.020} &
\meanstd{-0.322}{0.067} \\

FedProx &
\meanstd{0.141}{0.024} & \meanstd{0.113}{0.017} &
\meanstd{0.463}{0.096} & \meanstd{-0.172}{0.016} &
\meanstd{1.857}{0.761} & \meanstd{0.823}{0.000} &
\meanstd{0.478}{0.008} & \meanstd{-0.216}{0.011} &
\meanstd{-0.313}{0.081} \\

pFedMe &
\meanstd{0.324}{0.010} & \meanstd{0.290}{0.015} &
\meanstd{0.654}{0.040} & \meanstd{-0.054}{0.008} &
\meanstd{1.848}{0.270} & \meanstd{0.823}{0.000} &
\meanstd{0.477}{0.002} & \meanstd{-0.098}{0.013} &
\meanstd{-0.122}{0.026} \\

Ditto &
\meanstd{0.392}{0.022} & \meanstd{0.359}{0.029} &
\meanstd{0.685}{0.030} & \meanstd{-0.018}{0.007} &
\meanstd{1.697}{0.466} & \meanstd{0.825}{0.000} &
\meanstd{0.487}{0.006} & \meanstd{-0.062}{0.008} &
\meanstd{-0.091}{0.014} \\

FedDPA\textsuperscript{\dag} &
\meanstd{0.357}{0.035} & \meanstd{0.317}{0.033} &
\meanstd{0.667}{0.048} & \meanstd{-0.053}{0.021} &
\meanstd{1.511}{0.087} & \meanstd{0.825}{0.000} &
\meanstd{0.488}{0.005} & \meanstd{-0.097}{0.016} &
\meanstd{-0.110}{0.034} \\

FAVoR &
\meanstd{0.561}{0.024} & \meanstd{0.521}{0.029} &
\meanstd{0.771}{0.021} & \meanstd{0.093}{0.017} &
\meanstd{1.670}{0.304} & \meanstd{0.826}{0.000} &
\meanstd{0.494}{0.005} & \meanstd{0.049}{0.022} &
\meanstd{-0.006}{0.005} \\
\bottomrule
\end{tabularx}

\caption{Detailed Mythos-Reddit diagnostics.}
\label{tab:mythos-reddit-external-details}
\end{table*}

\subsection{FedDPA Gate Sensitivity Probe}
\label{subsec:feddpa_gate_sensitivity}

FedDPA stores prompt-wise dynamic gate references during training, while the main continuation path reports the exported per-client mean validation gate for generation compatibility.
To audit this export, we run a bounded seed probe on 100 matched BlogText held-out continuation prompts, generating one continuation with the exported summary gate and one with prompt-wise dynamic gates.
Prompt-wise dynamic gating changes the gate weights: the mean global-gate weight is 0.350 under dynamic gating versus 0.429 under the exported summary gate, with mean absolute difference 0.083, median absolute difference 0.056, and maximum absolute difference 0.437.
Table~\ref{tab:feddpa-gate-sensitivity} shows that this gate change leaves author accuracy unchanged and BERTScore nearly identical in the bounded probe.
Dynamic gating slightly lowers verification AUC and SemSim in this sample, so reporting the audited summary-gate FedDPA rows does not overstate FedDPA's author-style retention.

\begin{table}[H]
\centering
\footnotesize
\setlength{\tabcolsep}{3.2pt}
\renewcommand{\arraystretch}{1.06}
\begin{tabular}{lrrr}
\toprule
Metric & \makecell{Summary\\gate} & \makecell{Dynamic\\gate} & \makecell{Dynamic $-$\\summary} \\
\midrule
Author Acc. & 0.350 & 0.350 & +0.000 \\
Macro-F1 & 0.266 & 0.283 & +0.018 \\
Verif. AUC & 0.778 & 0.763 & -0.015 \\
EER & 0.299 & 0.303 & +0.004 \\
BERTScore & 0.8191 & 0.8178 & -0.0013 \\
SemSim & 0.3982 & 0.3827 & -0.0155 \\
Asst. score & 0.4035 & 0.4302 & +0.0267 \\
\bottomrule
\end{tabular}
\caption{Bounded FedDPA gate-sensitivity probe.}
\label{tab:feddpa-gate-sensitivity}
\end{table}

\subsection{Component-Ablation Significance and Run Checks}
\label{subsec:ablation_significance_audit}

Table~\ref{tab:phase1-ablation-full-meanstd} expands the compact main-text ablation table with mean $\pm$ sample standard deviation over three seeds.
Its Author Acc. and Macro-F1 columns are generated-text ASCE prototype metrics, while the StyleDistance, MiniLM, and stylometry columns are ASCE-independent, same-topic gen$\rightarrow$source AUC/EER.

\begin{table*}[t]
\centering
\footnotesize
\renewcommand{\arraystretch}{1.14}

\textit{Panel A: Component configuration and core metrics}\par\vspace{2pt}

\setlength{\tabcolsep}{1.7pt}
\begin{tabularx}{\textwidth}{
@{}>{\raggedright\arraybackslash}Xcccccccc@{}}
\toprule
Variant &
\makecell{Shared\\agg.} &
\makecell{Personal\\branch} &
\makecell{Private\\residual} &
\makecell{ASCE\\align.} &
\makecell{Author\\Acc. $\uparrow$} &
\makecell{Macro-\\F1 $\uparrow$} &
\makecell{Asst.\\Score $\downarrow$} &
\makecell{BERTScore\\$\uparrow$} \\
\midrule
Global ASCE &
\yesmark & \nomark & \nomark & \yesmark &
\meanstd{0.161}{0.014} &
\meanstd{0.140}{0.012} &
\meanstd{0.301}{0.065} &
\meanstd{\textbf{0.818}}{0.002} \\

Ditto + ASCE &
\yesmark & \yesmark & \nomark & \yesmark &
\meanstd{0.318}{0.037} &
\meanstd{0.286}{0.046} &
\meanstd{0.409}{0.040} &
\meanstd{0.816}{0.002} \\

FAVoR w/o ASCE &
\yesmark & \yesmark & \yesmark & \nomark &
\meanstd{0.377}{0.036} &
\meanstd{0.331}{0.033} &
\meanstd{0.306}{0.018} &
\meanstd{0.814}{0.001} \\

FAVoR &
\yesmark & \yesmark & \yesmark & \yesmark &
\meanstd{\textbf{0.403}}{0.004} &
\meanstd{\textbf{0.365}}{0.009} &
\meanstd{\textbf{0.260}}{0.004} &
\meanstd{0.815}{0.000} \\
\bottomrule
\end{tabularx}

\vspace{5pt}
\textit{Panel B: ASCE-independent external verification}\par\vspace{2pt}

\setlength{\tabcolsep}{7pt}
\begin{tabularx}{\textwidth}{
@{}>{\raggedright\arraybackslash}Xccc@{}}
\toprule
Variant &
\makecell{StyleDistance\\AUC/EER $\uparrow/\downarrow$} &
\makecell{MiniLM\\AUC/EER $\uparrow/\downarrow$} &
\makecell{Stylometry\\AUC/EER $\uparrow/\downarrow$} \\
\midrule
Global ASCE &
\meanstd{0.728/0.326}{0.009/0.003} &
\meanstd{0.638/0.407}{0.010/0.015} &
\meanstd{0.643/0.392}{0.009/0.016} \\

Ditto + ASCE &
\meanstd{0.746/0.319}{0.015/0.011} &
\meanstd{0.662/0.389}{0.005/0.006} &
\meanstd{0.707/0.345}{0.013/0.011} \\

FAVoR w/o ASCE &
\meanstd{\underline{0.767}/\underline{0.301}}{0.013/0.016} &
\meanstd{\underline{0.664}/\underline{0.388}}{0.003/0.007} &
\meanstd{\textbf{0.717}/\textbf{0.335}}{0.001/0.005} \\

FAVoR &
\meanstd{\textbf{0.771}/\textbf{0.295}}{0.006/0.003} &
\meanstd{\textbf{0.675}/\textbf{0.378}}{0.010/0.006} &
\meanstd{\underline{0.714}/\underline{0.342}}{0.011/0.011} \\
\bottomrule
\end{tabularx}

\caption{Full component-ablation summary with seed variability.
Panel A reports component configurations and core BlogText metrics;
Panel B reports ASCE-independent same-topic
gen$\rightarrow$source AUC/EER. Author Acc. and Macro-F1 are
generated-text ASCE prototype metrics. Values are mean $\pm$
sample standard deviation over three seeds.}
\label{tab:phase1-ablation-full-meanstd}
\end{table*}

Table~\ref{tab:phase1-ablation-paired-tests} reports paired tests for the component ablations in Table~\ref{tab:phase1-main-results}.
The tests use matched seed--author--prompt examples.
For author accuracy, $p$ uses McNemar's exact test; for continuous metrics, $p$ uses a Wilcoxon signed-rank test.
Directional $\Delta$ is signed, so positive values favour the left variant in each comparison; for the assistant score, this means the left variant has a lower raw assistant score.
Confidence intervals are cluster bootstrap intervals over seed--author clusters.

\begin{table*}[t]
\centering
\footnotesize
\setlength{\tabcolsep}{3.2pt}
\renewcommand{\arraystretch}{1.05}
\begin{tabular}{llrrrrr}
\toprule
Comparison & Metric & Left & Right & \makecell{Directional\\$\Delta$} & \makecell{Cluster 95\%\\CI} & $p$ \\
\midrule
FAVoR vs. w/o ASCE & Author acc. & 0.403 & 0.377 & +0.026 & [-0.012, 0.064] & 0.225 \\
FAVoR vs. w/o ASCE & Top-1 cosine & 0.642 & 0.629 & +0.013 & [0.003, 0.023] & 0.011 \\
FAVoR vs. w/o ASCE & Asst. score & 0.260 & 0.306 & +0.046 & [0.012, 0.081] & 0.009 \\
FAVoR vs. w/o ASCE & BERTScore & 0.815 & 0.814 & +0.001 & [-0.001, 0.002] & 0.410 \\
FAVoR vs. w/o ASCE & SemSim & 0.380 & 0.377 & +0.003 & [-0.006, 0.012] & 0.248 \\
\midrule
FAVoR vs. Global ASCE & Author acc. & 0.403 & 0.161 & +0.242 & [0.194, 0.291] & $<.001$ \\
FAVoR vs. Global ASCE & Top-1 cosine & 0.642 & 0.678 & -0.036 & [-0.052, -0.020] & $<.001$ \\
FAVoR vs. Global ASCE & Asst. score & 0.260 & 0.301 & +0.041 & [-0.004, 0.085] & 0.038 \\
FAVoR vs. Global ASCE & BERTScore & 0.815 & 0.818 & -0.004 & [-0.006, -0.002] & 0.004 \\
FAVoR vs. Global ASCE & SemSim & 0.380 & 0.364 & +0.017 & [0.006, 0.027] & 0.001 \\
\midrule
w/o ASCE vs. Global ASCE & Author acc. & 0.377 & 0.161 & +0.216 & [0.170, 0.263] & $<.001$ \\
w/o ASCE vs. Global ASCE & Top-1 cosine & 0.629 & 0.678 & -0.049 & [-0.067, -0.031] & $<.001$ \\
w/o ASCE vs. Global ASCE & Asst. score & 0.306 & 0.301 & -0.005 & [-0.055, 0.043] & 0.874 \\
w/o ASCE vs. Global ASCE & BERTScore & 0.814 & 0.818 & -0.004 & [-0.006, -0.002] & 0.002 \\
w/o ASCE vs. Global ASCE & SemSim & 0.377 & 0.364 & +0.013 & [0.002, 0.024] & 0.016 \\
\bottomrule
\end{tabular}
\caption{Paired component-ablation significance tests.}
\label{tab:phase1-ablation-paired-tests}
\end{table*}

Table~\ref{tab:phase1-ablation-external-paired} repeats the ablation comparison using ASCE-independent external gen$\rightarrow$source verification.
Under the same-topic cross-author negatives, the external alignment gain over the no-ASCE variant is small and mixed, while both private-residual variants are consistently stronger than the Global ASCE control.
Directional $\Delta$ is signed so positive values favour the left variant; for EER, this means lower EER for the left variant. Finally, Table~\ref{tab:phase1-ablation-run-audit} reports a compact configuration check for the ablation rows.
It verifies whether each run uses a private residual, ASCE alignment, and the shared proximal term.
For held-out continuation, the deterministic quality filter removes generations with fewer than 20 whitespace-delimited tokens, including empty generations, or a distinct-bigram ratio below $0.15$.

\begin{table*}[t]
\centering
\scriptsize
\setlength{\tabcolsep}{4pt}
\resizebox{\textwidth}{!}{%
\begin{tabular}{lllrrrrr}
\toprule
Comparison & Evaluator & Metric & Left & Right & \makecell{Directional\\$\Delta$} & \makecell{Cluster 95\%\\CI} & $p$ \\
\midrule
FAVoR vs. w/o ASCE & StyleDistance & AUC & 0.771 & 0.767 & +0.004 & [-0.007, 0.015] & 0.498 \\
FAVoR vs. w/o ASCE & StyleDistance & EER & 0.295 & 0.301 & +0.006 & [-0.007, 0.019] & 0.438 \\
FAVoR vs. w/o ASCE & MiniLM & AUC & 0.675 & 0.664 & +0.011 & [-0.002, 0.025] & 0.102 \\
FAVoR vs. w/o ASCE & MiniLM & EER & 0.378 & 0.388 & +0.010 & [-0.003, 0.025] & 0.129 \\
FAVoR vs. w/o ASCE & Stylometry & AUC & 0.714 & 0.717 & -0.003 & [-0.013, 0.006] & 0.492 \\
FAVoR vs. w/o ASCE & Stylometry & EER & 0.342 & 0.334 & -0.007 & [-0.019, 0.005] & 0.233 \\
\midrule
FAVoR vs. Global ASCE & StyleDistance & AUC & 0.771 & 0.728 & +0.043 & [0.026, 0.061] & $<.001$ \\
FAVoR vs. Global ASCE & StyleDistance & EER & 0.295 & 0.327 & +0.032 & [0.014, 0.048] & $<.001$ \\
FAVoR vs. Global ASCE & MiniLM & AUC & 0.675 & 0.638 & +0.037 & [0.021, 0.053] & $<.001$ \\
FAVoR vs. Global ASCE & MiniLM & EER & 0.378 & 0.408 & +0.030 & [0.015, 0.046] & $<.001$ \\
FAVoR vs. Global ASCE & Stylometry & AUC & 0.714 & 0.643 & +0.071 & [0.049, 0.094] & $<.001$ \\
FAVoR vs. Global ASCE & Stylometry & EER & 0.342 & 0.392 & +0.050 & [0.031, 0.072] & $<.001$ \\
\midrule
w/o ASCE vs. Global ASCE & StyleDistance & AUC & 0.767 & 0.728 & +0.039 & [0.023, 0.056] & $<.001$ \\
w/o ASCE vs. Global ASCE & StyleDistance & EER & 0.301 & 0.327 & +0.025 & [0.009, 0.044] & 0.003 \\
w/o ASCE vs. Global ASCE & MiniLM & AUC & 0.664 & 0.638 & +0.026 & [0.009, 0.043] & 0.004 \\
w/o ASCE vs. Global ASCE & MiniLM & EER & 0.388 & 0.408 & +0.020 & [0.004, 0.035] & 0.011 \\
w/o ASCE vs. Global ASCE & Stylometry & AUC & 0.717 & 0.643 & +0.074 & [0.054, 0.096] & $<.001$ \\
w/o ASCE vs. Global ASCE & Stylometry & EER & 0.334 & 0.392 & +0.057 & [0.040, 0.077] & $<.001$ \\
\bottomrule
\end{tabular}%
}
\caption{External-verifier paired tests for component ablations using ASCE-independent same-topic gen$\rightarrow$source verification. Directional $\Delta$ is signed so positive values favor the left variant; for EER, positive directional $\Delta$ means lower raw EER for the left variant.}
\label{tab:phase1-ablation-external-paired}
\end{table*}

\begin{table}[H]
\centering
\small
\setlength{\tabcolsep}{5pt}
\renewcommand{\arraystretch}{1.10}

\begin{tabular}{@{}lccc@{}}
\toprule
Variant &
\makecell{Private\\residual} &
\makecell{ASCE\\align.} &
\makecell{Shared\\prox.} \\
\midrule
Global ASCE    & \nomark & \yesmark & \nomark \\
FAVoR w/o ASCE & \yesmark & \nomark & \yesmark \\
FAVoR          & \yesmark & \yesmark & \yesmark \\
\bottomrule
\end{tabular}

\caption{Component-ablation configuration check.}
\label{tab:phase1-ablation-run-audit}
\end{table}

\subsection{Hyperparameters, Efficiency, and Communication Cost}
\label{subsec:experiment_efficiency}

Table~\ref{tab:efficiency_communication} summarizes the hyperparameters and the communication and storage footprint of the BlogText FAVoR runs.
The measurements are taken from the three medium non-IID FAVoR runs used in the main BlogText evaluation.
All runs use Qwen2.5-3B-Instruct with 4-bit Quantized Low-Rank Adaptation (QLoRA) \citep{qwen25-3b-instruct,dettmers-etal-2023-qlora} and train LoRA adapters on the $q$, $k$, $v$, $o$, gate, up, and down projections with rank $16$, $\alpha=32$, and dropout $0.05$.
The serialized shared adapter is 119{,}801{,}528 bytes, corresponding to 114.3 MiB and 29.93M trainable LoRA parameters.
These settings were set before the held-out test evaluation; no hyperparameter was tuned on test continuations.
Under the protocol, each sampled client uploads only the shared-adapter delta.
With 12 clients per round for six rounds, one seed sends 72 client updates, or 8.03 GiB of client-to-server adapter payloads when counted as serialized fp32 LoRA tensors.
Server-to-client broadcasts have the same per-recipient size, so counting both directions gives 16.07 GiB per seed.
Private residual packs are not communicated; retaining one final residual per 50-author evaluation roster would require 5.58 GiB if stored as dense fp32 LoRA factor-delta tensors.

\begin{table}[H]
\centering
\small
\begin{tabularx}{\linewidth}{@{}>{\raggedright\arraybackslash}p{0.30\linewidth}>{\raggedright\arraybackslash}X@{}}
\toprule
Quantity & Value \\
\midrule
Base model & Qwen2.5-3B-Instruct \\
Quantization & 4-bit QLoRA, NF4, bfloat16 compute \\
Trainable PEFT params & 29.93M \\
Serialized shared adapter & 114.3 MiB \\
Rounds & 6 \\
Clients per round & 12 of 50 \\
Local epochs & 1 shared, 2 private \\
Batching & micro-batch 4, grad. accumulation 8 \\
Learning rate & $2 \times 10^{-4}$ shared/private \\
Upload per client round & 114.3 MiB \\
Aggregate upload per round & 1.34 GiB \\
Aggregate upload per seed & 8.03 GiB \\
Upload + broadcast per seed & 16.07 GiB \\
Final private residual storage & 5.58 GiB for 50 clients \\
Runtime per seed & 4.40--4.60 hours on 4 RTX 4090 GPUs \\
\bottomrule
\end{tabularx}
\caption{Efficiency and communication cost for the main BlogText FAVoR runs.}
\label{tab:efficiency_communication}
\end{table}

\subsection{Statistical Testing}
\label{subsec:experiment_significance}

Table~\ref{tab:phase1-paired-significance} reports paired evidence for the main BlogText comparisons over matched seed--author--prompt examples.
Table~\ref{tab:phase1-external-paired-auc} extends the same question to ASCE-independent external gen$\rightarrow$source verification by recomputing AUC and EER under seed--author cluster bootstrap resampling.
The paired evidence supports the main interpretation: FAVoR's author-style improvements over Ditto, pooled PEFT, FedProx, and FedDPA are robust, while semantic utility differences are smaller and not uniformly in FAVoR's favor.
Because each comparison is restricted to the method-pair intersection of quality-passed examples, the FAVoR and baseline columns are recomputed within each comparison block and need not exactly match the corresponding full-coverage summary-table means.
In these paired-test tables, directional $\Delta$ is signed so positive values favour FAVoR; for assistant score and EER, this means lower raw values for FAVoR.
For Table~\ref{tab:phase1-paired-significance}, binary author-accuracy $p$ uses McNemar's exact test, while continuous BlogText metrics use the same seed--author cluster-bootstrap distribution for confidence intervals and two-sided tail-probability $p$ values. For Table~\ref{tab:phase1-external-paired-auc}, AUC/EER confidence intervals and $p$ values likewise both come from the same two-sided cluster-bootstrap tail probability.

\makeatletter
\setlength{\@dblfptop}{0pt}
\setlength{\@dblfpsep}{2pt}
\setlength{\@dblfpbot}{0pt plus 1fil}
\makeatother
\setlength{\belowcaptionskip}{0pt}

\begin{table*}[t]
\centering
\footnotesize
\setlength{\tabcolsep}{3.2pt}
\renewcommand{\arraystretch}{1.04}
\begin{tabular}{llrrrrcr}
\toprule
Comparison & Metric & FAVoR & Baseline & \makecell{Directional\\$\Delta$} & \makecell{Cluster 95\%\\CI} & $p$ & $n$ \\
\midrule
FAVoR vs. Ditto & Author acc. & 0.403 & 0.279 & +0.125 & [0.078, 0.170] & $<.001$ & 578 \\
FAVoR vs. Ditto & Top-1 cosine & 0.642 & 0.655 & -0.013 & [-0.026, -0.001] & 0.037 & 578 \\
FAVoR vs. Ditto & SemSim & 0.380 & 0.366 & +0.014 & [0.004, 0.024] & 0.004 & 578 \\
FAVoR vs. Ditto & BERTScore & 0.815 & 0.815 & -0.001 & [-0.002, 0.001] & 0.405 & 578 \\
FAVoR vs. Ditto & Asst. score & 0.260 & 0.328 & +0.068 & [0.023, 0.114] & 0.002 & 578 \\
\midrule
FAVoR vs. Pooled & Author acc. & 0.408 & 0.255 & +0.153 & [0.110, 0.198] & $<.001$ & 569 \\
FAVoR vs. Pooled & Top-1 cosine & 0.643 & 0.590 & +0.052 & [0.036, 0.069] & $<.001$ & 569 \\
FAVoR vs. Pooled & SemSim & 0.383 & 0.359 & +0.024 & [0.013, 0.036] & $<.001$ & 569 \\
FAVoR vs. Pooled & BERTScore & 0.815 & 0.802 & +0.013 & [0.010, 0.015] & $<.001$ & 569 \\
FAVoR vs. Pooled & Asst. score & 0.261 & 0.219 & -0.041 & [-0.089, 0.006] & 0.083 & 569 \\
\midrule
FAVoR vs. FedProx & Author acc. & 0.403 & 0.175 & +0.228 & [0.182, 0.276] & $<.001$ & 578 \\
FAVoR vs. FedProx & Top-1 cosine & 0.642 & 0.671 & -0.029 & [-0.046, -0.013] & $<.001$ & 578 \\
FAVoR vs. FedProx & SemSim & 0.380 & 0.353 & +0.027 & [0.016, 0.038] & $<.001$ & 578 \\
FAVoR vs. FedProx & BERTScore & 0.815 & 0.817 & -0.002 & [-0.004, -0.000] & 0.024 & 578 \\
FAVoR vs. FedProx & Asst. score & 0.260 & 0.264 & +0.004 & [-0.041, 0.050] & 0.857 & 578 \\
\midrule
FAVoR vs. FedDPA & Author acc. & 0.403 & 0.294 & +0.109 & [0.064, 0.155] & $<.001$ & 578 \\
FAVoR vs. FedDPA & Top-1 cosine & 0.642 & 0.684 & -0.042 & [-0.057, -0.028] & $<.001$ & 578 \\
FAVoR vs. FedDPA & SemSim & 0.380 & 0.376 & +0.004 & [-0.006, 0.014] & 0.400 & 578 \\
FAVoR vs. FedDPA & BERTScore & 0.815 & 0.818 & -0.004 & [-0.006, -0.002] & $<.001$ & 578 \\
FAVoR vs. FedDPA & Asst. score & 0.260 & 0.419 & +0.159 & [0.108, 0.211] & $<.001$ & 578 \\
\bottomrule
\end{tabular}
\caption{Paired tests over matched seed--author--prompt examples. FAVoR and baseline values are recomputed on each method-pair matched subset, so the same method can have slightly different displayed values across comparison blocks and need not exactly equal its full-coverage summary-table mean. Directional $\Delta$ is signed, so positive values favour FAVoR; for the assistant score, lower raw values are better. Confidence intervals are cluster-bootstrap intervals over seed--author clusters. For author accuracy, $p$ uses McNemar's exact test; for continuous metrics, $p$ uses the two-sided cluster-bootstrap tail probability.}
\label{tab:phase1-paired-significance}
\end{table*}

\begin{table*}[t]
\centering
\footnotesize
\setlength{\tabcolsep}{3pt}
\renewcommand{\arraystretch}{1.04}
\begin{tabular}{lllrrrrr}
\toprule
Comparison & Evaluator & Metric & FAVoR & Baseline & \makecell{Directional\\$\Delta$} & \makecell{Cluster 95\%\\CI} & $p$ \\
\midrule
FAVoR vs. Ditto & StyleDistance & AUC & 0.771 & 0.748 & +0.023 & [0.009, 0.038] & 0.001 \\
FAVoR vs. Ditto & StyleDistance & EER & 0.295 & 0.319 & +0.024 & [0.008, 0.037] & 0.002 \\
FAVoR vs. Ditto & MiniLM & AUC & 0.675 & 0.650 & +0.025 & [0.009, 0.042] & 0.004 \\
FAVoR vs. Ditto & MiniLM & EER & 0.378 & 0.402 & +0.024 & [0.005, 0.040] & 0.012 \\
FAVoR vs. Ditto & Stylometry & AUC & 0.714 & 0.683 & +0.031 & [0.009, 0.055] & 0.004 \\
FAVoR vs. Ditto & Stylometry & EER & 0.342 & 0.362 & +0.021 & [0.002, 0.043] & 0.026 \\
\midrule
FAVoR vs. Pooled & StyleDistance & AUC & 0.771 & 0.736 & +0.035 & [0.020, 0.050] & $<.001$ \\
FAVoR vs. Pooled & StyleDistance & EER & 0.295 & 0.327 & +0.032 & [0.016, 0.049] & $<.001$ \\
FAVoR vs. Pooled & MiniLM & AUC & 0.675 & 0.649 & +0.026 & [0.007, 0.045] & 0.005 \\
FAVoR vs. Pooled & MiniLM & EER & 0.378 & 0.406 & +0.028 & [0.010, 0.046] & 0.001 \\
FAVoR vs. Pooled & Stylometry & AUC & 0.714 & 0.660 & +0.054 & [0.034, 0.076] & $<.001$ \\
FAVoR vs. Pooled & Stylometry & EER & 0.342 & 0.384 & +0.043 & [0.024, 0.064] & $<.001$ \\
\midrule
FAVoR vs. FedProx & StyleDistance & AUC & 0.771 & 0.723 & +0.047 & [0.031, 0.065] & $<.001$ \\
FAVoR vs. FedProx & StyleDistance & EER & 0.295 & 0.334 & +0.039 & [0.020, 0.055] & $<.001$ \\
FAVoR vs. FedProx & MiniLM & AUC & 0.675 & 0.631 & +0.045 & [0.028, 0.061] & $<.001$ \\
FAVoR vs. FedProx & MiniLM & EER & 0.378 & 0.414 & +0.036 & [0.021, 0.052] & $<.001$ \\
FAVoR vs. FedProx & Stylometry & AUC & 0.714 & 0.650 & +0.063 & [0.042, 0.086] & $<.001$ \\
FAVoR vs. FedProx & Stylometry & EER & 0.342 & 0.385 & +0.043 & [0.025, 0.066] & $<.001$ \\
\midrule
FAVoR vs. FedDPA & StyleDistance & AUC & 0.771 & 0.748 & +0.023 & [0.009, 0.037] & $<.001$ \\
FAVoR vs. FedDPA & StyleDistance & EER & 0.295 & 0.316 & +0.021 & [0.006, 0.036] & 0.004 \\
FAVoR vs. FedDPA & MiniLM & AUC & 0.675 & 0.659 & +0.016 & [0.001, 0.031] & 0.035 \\
FAVoR vs. FedDPA & MiniLM & EER & 0.378 & 0.391 & +0.013 & [0.000, 0.029] & 0.049 \\
FAVoR vs. FedDPA & Stylometry & AUC & 0.714 & 0.692 & +0.022 & [0.005, 0.038] & 0.008 \\
FAVoR vs. FedDPA & Stylometry & EER & 0.342 & 0.351 & +0.009 & [-0.005, 0.026] & 0.201 \\
\bottomrule
\end{tabular}
\caption{External-verifier paired tests for main BlogText comparisons.}
\label{tab:phase1-external-paired-auc}
\end{table*}
\end{document}